\documentclass[letterpaper]{article} 
\usepackage{aaai2027}  
\usepackage[hyphens]{url}  
\usepackage{graphicx} 
\usepackage{natbib}  
\usepackage{caption} 
\usepackage{algorithm}
\usepackage{algorithmic}
\usepackage{subcaption}
\usepackage{newfloat}
\usepackage{listings}
\DeclareCaptionStyle{ruled}{labelfont=normalfont,labelsep=colon,strut=off} 
\floatstyle{ruled}
\newfloat{listing}{tb}{lst}{}
\floatname{listing}{Listing}

\usepackage{booktabs}

\usepackage{amsmath,amssymb,amsthm}
\usepackage{multirow}
\usepackage{tikz}
\usetikzlibrary{arrows.meta,positioning,fit,calc}

\newcommand{\ie}{i.e.,\ }

\newcommand{\xmask}{x'}
\newcommand{\TT}{\mathsf{T}}
\newcommand{\method}{\textsc{GRAIL}}
\newcommand{\amort}{\textsc{GRAIL-fast}}
\newcommand{\auto}{\textsc{GRAIL-auto}}
\newcommand{\best}[1]{\textbf{#1}}
\newcommand{\runner}[1]{\underline{#1}}

\newcommand{\first}{\emph{(i)~}}
\newcommand{\second}{\emph{(ii)~}}
\newcommand{\third}{\emph{(iii)~}}

\usepackage{xcolor}
\usepackage{soul}
\sethlcolor{yellow}  

\theoremstyle{definition}
\newtheorem{example}{Example}

\title{GRAIL: From Saliency to Semantics in Time-Series Explanation}

\author {
    Enver Menadjiev\textsuperscript{\rm 1}\corresponding,
    Yiqiu Shen\textsuperscript{\rm 2},
    Jaesik Choi\textsuperscript{\rm 1},
    Julia Stoyanovich\textsuperscript{\rm 2}\corresponding
}
\affiliations {
    \textsuperscript{\rm 1}KAIST\\
    \textsuperscript{\rm 2}NYU\\
    enver@kaist.ac.kr, ys1001@nyu.edu, jaesik.choi@kaist.ac.kr, stoyanovich@nyu.edu
}

\begin{document}

\maketitle

\begin{abstract}
Explaining a time-series model requires both locating the evidence and extracting meaningful knowledge from it. 
Yet a single timestep is not a semantic unit: the meaningful patterns live across many samples, not in any one of them. Per-timestep saliency is therefore noisy, high-dimensional, and unstable. Existing methods address this by introducing fixed-size windows and treating them as independent players in a cooperative game. Evaluating a coalition of such players requires defining a ``baseline'' region, which prior work tackles with fixed counterfactuals (e.g., a flat line, noise, or a spectral surrogate). This choice frequently dominates the resulting explanation, obscuring the model's true decisions. 

To address these issues, we propose \method{}, a region-based time-series explanation framework that uses domain-specific intervals, such as the P-wave and the QRS complex of an electrocardiogram (ECG) beat, as players. We describe each region with human-readable features and generate counterfactual baselines with an on-manifold conditional generator. Such a choice of players also admits non-temporal features: static covariates enter as singleton players attributed with the temporal regions on one scale, so a patient's tabular evidence (e.g., age, sex) is weighed directly against the signal. We attribute the predictions to the features via Owen values with common-random-number variance reduction. Experimental results on both synthetic and real data show that the proposed framework produces faithful and robust explanations, evaluated under matched on- and off-manifold operators, offering meaningful knowledge to domain experts.
\end{abstract}
\section{Introduction}\label{sec:intro}

Deep learning models are increasingly deployed on high-stakes time series, from clinical data to industrial sensors and financial data. Wherever they are used, deploying them hinges on a question that's hard to answer for time series: which part of the signal drove a decision, in terms a domain expert can act on. 


\begin{figure}[t]
\centering
\begin{subfigure}{\columnwidth}
  \centering
  \includegraphics[width=\columnwidth]{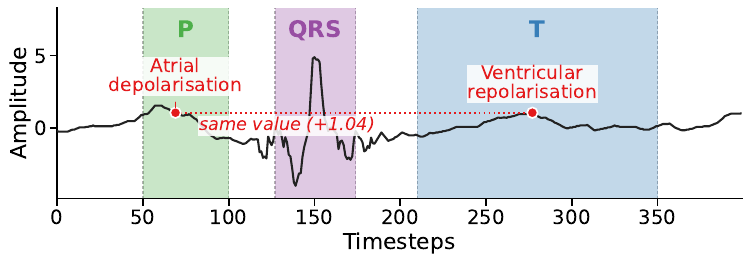}
  \caption{a cardiologist-delineated ECG beat}\label{fig:teaser:signal}
\end{subfigure}
\begin{subfigure}{\columnwidth}
  \centering
  \includegraphics[width=\columnwidth]{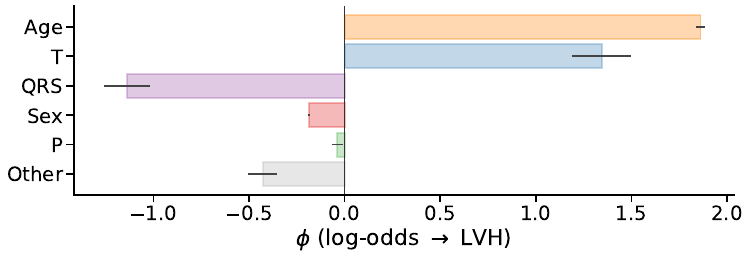}
  \caption{an explanation of the LVH-positive prediction by \method{}}\label{fig:teaser:phi}
\end{subfigure}

\caption{\textbf{From saliency to semantics on one LUDB beat.} (a) A real ECG beat split into P, QRS, T, and the isoelectric background. Two samples of equal amplitude ($+1.04$) fall in different regions, so a per-timestep view cannot separate them. (b) \method{} scores each region and static covariate (age, sex) on a single log-odds scale for this beat, which the model correctly classifies as LVH-absent. The prediction relies primarily on the QRS region, even though the patient's age and T region would suggest otherwise.}
\label{fig:teaser}
\end{figure}

\begin{example}\label{intro:ex}
Consider a cardiologist inspecting a single heartbeat (Figure~\ref{fig:teaser:signal}). Rather than reasoning sample by sample, the cardiologist reads a sequence of named regions: the P-wave (atrial activation), the QRS complex (the sharp ventricular spike), and the T-wave (ventricular recovery), and attaches clinical meaning to each. Left-ventricular hypertrophy (LVH), for instance, is read chiefly from the QRS voltage. A saliency map that reports ``sample 137 was important'' may localize evidence, but it does not directly express the regional pattern or its clinical meaning.  
\end{example}

\begin{figure*}[t]
\centering
\begin{tikzpicture}[
  font=\footnotesize,
  box/.style={draw, rounded corners, align=center, minimum height=9mm, inner sep=3pt, fill=white},
  op/.style={draw, rounded corners, align=center, minimum height=9mm, inner sep=3pt, fill=black!5},
  fast/.style={draw, rounded corners, align=center, minimum height=9mm, inner sep=3pt, fill=blue!6},
  col/.style={text width=23mm, minimum height=9mm},
  arr/.style={-{Latex[length=2mm]}, thick},
]
  \node[box] (x) {signal $x$};
  \node[op, right=6mm of x] (seg) {\textbf{Segmenter}\\ delineation /\\ model-aware};
  \node[box, right=6mm of seg] (reg) {regions\\ $R_1,\dots,R_N$};
  \node[op, right=6mm of reg] (enc) {\textbf{Codec}\\ named features\\ $\hat\theta$ per region};
  \node[op, right=6mm of enc] (game) {\textbf{Game} $v(S)$\\ absent $\Rightarrow$\\ \emph{regenerate}\\ on-manifold};

  \coordinate (branch) at ([xshift=6mm]game.east);
  \node[op, col, anchor=south west] (est) at ([yshift=1mm]branch)
     {\textbf{Owen+CRN}\\ estimator\\ (exact)};
  \node[fast, col, anchor=north west] (amort) at ([yshift=-1mm]branch)
     {\textbf{Amortized net}\\ $g_\psi(x)$ trained on the game};

  \coordinate (colr) at (est.east |- x);
  \node[box, anchor=west] (out) at ([xshift=6mm]colr) {$\phi$ + region\\ roll-up + $\rho$};

  \draw[arr] (x) -- (seg);  \draw[arr] (seg) -- (reg); \draw[arr] (reg) -- (enc);
  \draw[arr] (enc) -- (game);
  \draw[arr] (game.east) -- (est.west); \draw[arr] (est.east) -- (out.west);
  \draw[arr, blue!55] (game.east) -- (amort.west);
  \draw[arr, blue!55] (amort.east) -- (out.west);
\end{tikzpicture}
\caption{\textbf{The \method{} pipeline.} A signal is segmented into named regions; a \emph{codec} encodes each as human-readable features and can regenerate it. In the game, a \emph{present} feature keeps its observed value while an \emph{absent} one triggers \emph{on-manifold regeneration}, so every coalition is a plausible real signal. Owen values (features nested in regions) are estimated exactly with a common-random-number sweep, yielding a complete, region-rolled-up attribution plus the residual $\rho$. An \emph{amortized} network (blue), trained on the same game, reproduces the attribution in one forward pass for fast, robust deployment.}
\label{fig:schematic}
\end{figure*}
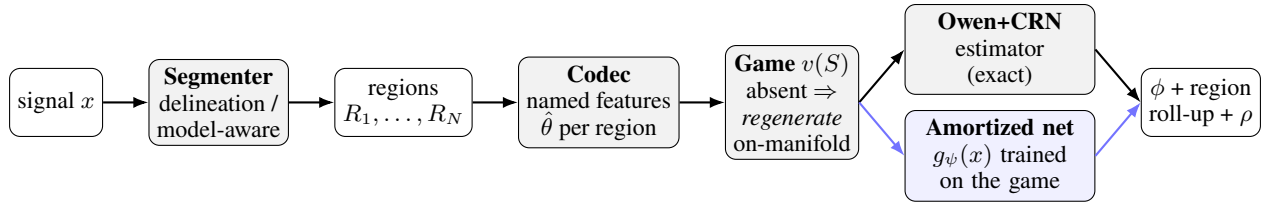


Predictive patterns typically span multiple samples, rendering per-timestep saliency maps high-dimensional, unstable, and difficult for domain experts to interpret~\cite{ismail2020benchmarking}. This motivates our first design choice: explain in the domain expert's units. We take the unit of explanation to be the region, a contiguous, named interval corresponding to a recognizable event: the P/QRS/T waves of an ECG beat, or an injected pulse, a seasonal stretch, or a trend in a synthetic signal. Regions are few, human-readable, and carry a prior, since cardiology tells us which region should distinguish an LVH from a normal beat.

Additionally, models are routinely given non-temporal information alongside the temporal signal. Returning to Example~\ref{intro:ex}, a cardiologist reads an ECG differently depending on the patient's sex and age, since the voltage criteria for left-ventricular hypertrophy are sex-specific and normal intervals shift with age. This motivates our second design choice: to place the signal and covariates on a common scale. A faithful explanation must weigh such covariates against the waveform, so a model's reliance on demographics in addition to morphology can be read directly from the model. Figure~\ref{fig:teaser:phi} provides a clinically meaningful explanation of the correct diagnosis: the model calls this 68-year-old LVH-positive despite a normal QRS, consistent with the lower QRS-voltage threshold that LVH criteria apply at older ages.

\paragraph{Our approach.}
We propose \method{} (Generative Region Attribution with Interpretable Labels), a framework that explains time-series predictions in domain-aligned units.  \method{} produces \emph{feature-based, additive} explanations via a cooperative game: each region and each covariate is a \emph{player}, and its value is its contribution to the prediction across all coalitions, so the contributions sum to the model's output. Scoring a coalition requires saying what the signal looks like when a region is \emph{absent}, but there is no canonical ``off'' state for a stretch of a time series such as an ECG: prior work fills it with a fixed counterfactual, a flat line, noise, or a spectrally matched surrogate, an arbitrary choice that frequently dominates the resulting explanation~\cite{sundararajan2020many}. The choice of the absent state is thus not an implementation detail; it \emph{is} the explanation. In this work, we stop choosing from a fixed menu and instead \emph{build} the counterfactual: an interpretable \emph{codec} encodes each region as named features, and a conditional \emph{generator} regenerates an absent region \emph{on the data manifold}, so every coalition remains a plausible real signal.

\method{} attributes the model's log-odds to regions and covariates with \emph{Owen values}~\cite{owen1972multilinear} (a two-level Shapley value~\cite{shapley1953value}, features nested in regions) estimated by a common-random-number sweep, and always reports the residual $\rho$, so the accounting is exactly complete. Figure~\ref{fig:schematic} shows the pipeline of \method{}, which also amortizes into a fast single-pass surrogate for deployment.

\paragraph{Contributions.}
\begin{enumerate}
    \item \textbf{A region-and-covariate cooperative game.} We cast time-series attribution as a cooperative game whose players are semantic temporal regions \emph{and} non-temporal covariates, credited on one common scale, so reliance on metadata is read off alongside the waveform. To our knowledge, no prior time-series explainer accommodates both (Section~\ref{sec:define}).
    \item \textbf{An exact, on-manifold attribution.} \method{} plays the game by \emph{generating} each absent region on the data manifold rather than from a fixed filler, and attributes with exactly-complete Owen values (CRN-estimated). This is the faithful but slow reference; a model-aware segmenter recovers regions absent a prior (Section~\ref{sec:play}).
    \item \textbf{A fast surrogate for deployment.} \method{}-fast amortizes the game into one forward pass, $\sim$$10^3\times$ cheaper and still exactly complete, though its on-manifold faithfulness holds only on real data
    (Section~\ref{sec:opt}).
    \item \textbf{A comprehensive benchmark.} An at-scale comparison over four families of benchmarks, including real ECG, scoring every method under \emph{matched} on- and off-manifold operators with paired significance tests against concept- and group-level Shapley baselines, with kernel, generator, reference, and segmenter ablations (Section~\ref{sec:exp}).
\end{enumerate}

\section{Related Work}\label{sec:related}

\paragraph{Attribution and time-series explainers.} Attribution via Shapley values~\cite{shapley1953value,lundberg2017unified} depends on how an ``absent'' feature is modeled~\cite{sundararajan2020many,janzing2020feature,covert2021explaining}. Most time-series explainers give per-timestep saliency~\cite{theissler2022explainable}; those at a coarser grain group timesteps into windows or segments (WindowSHAP~\cite{nayebi2023windowshap}, TimeSHAP~\cite{bento2021timeshap}, TS-MULE~\cite{schlegel2021tsmule}, LIMESegment~\cite{sivill2022limesegment}, Dynamask~\cite{crabb2021dynamask}), but all fill the absent state with a fixed, off-manifold filler. Instead, we make the removal operator a conditional generator that stays on-manifold and lift the flat Shapley game to a two-level Owen value~\cite{owen1972multilinear} over features nested within regions.

\paragraph{Closest comparators and evaluation.} Nearest to us are methods that highlight time series players: C-SHAP~\cite{jutte2025cshap} uses signal concepts, while GroupSegment-SHAP~\cite{kim2026gsshap}. We also benchmark these approaches. They redefine the player but keep an off-manifold mean-replacement operator, whereas we hold the player structure fixed and change the operator, and neither admits non-temporal covariates. CausalConceptTS~\cite{lopezalcaraz2025causalconcept} shares our generative premise but imputes over full recordings, and ManifoldShap~\cite{taufiq2023manifold} restricts the interventional background to an on-manifold region rather than generating it. Our per-region generation keeps an exact Owen sweep affordable. Because perturbation faithfulness is confounded by the removal operator~\cite{hooker2019benchmark,jethani2021have}, we score every method under both a matched on-manifold and a neutral off-manifold operator, and amortize the exact game for deployment~\cite{jethani2021fastshap}. Appendix~\ref{app:related} expands each baseline and the hierarchical-credit connection.
\section{The \method{} framework}\label{sec:method}

\method{} represents time-series attribution as a cooperative game and solves it on the data manifold. We define the game and its players, semantic regions together with non-temporal covariates in Section~\ref{sec:define}; play it exactly, generating each absent region's counterfactual on-manifold and attributing with exactly-complete Owen values in Section~\ref{sec:play}; and amortize the exact game into a fast single-pass surrogate for deployment in Section~\ref{sec:opt}.

\subsection{Defining the Cooperative Game}\label{sec:define}

We write a signal as $x\in\mathbb{R}^{C\times T}$ over $C$ channels ($C=1$ for a univariate series) and denote the model's logit for the target class by $f$. A segmenter partitions $x$ into named regions $R_1,\dots, R_M$, each keyed by (channel, span), and a codec (an interpretable encoder paired with a conditional generator) encodes each region as human-readable features $\hat\theta_r$. The players $N$ are the region features, grouped into Owen coalition blocks $B$. Static covariates are treated as singleton players appended after the regions, so that signal and tabular evidence are evaluated on a single scale. This ensures no contribution leakage into $v(\emptyset)$.

For a coalition $S\subseteq N$ we pin each present feature to its observed value and draw each absent feature from a reference,
\begin{equation}
\theta_i(S,\omega)=
\begin{cases}
    \hat\theta_i, & i\in S \quad(\text{present: observed}),\\
    F^{-1}_{p_{0,i}}(u_i), & i\notin S \quad(\text{absent: generated}),
\end{cases}
\end{equation}
with $u_i\sim\mathrm{Unif}[0,1]$ and $F^{-1}_{p_{0,i}}$ the quantile function of feature $i$'s reference marginal (App.~\ref{app:impl}). Each region is then regenerated from its features by a conditional generator, conditioned only on its own channel since boundary continuity is a within-channel property. Folding the reference draws $u$ and generator noise $\varepsilon$ into a fixed base randomness $\omega=(\varepsilon,u)$ makes the assembled signal a deterministic function $\xmask(S,\omega)$, and the coalition value is the expected logit
\begin{equation}
    v(S)=\mathbb{E}_{\omega}\!\big[\,f(\xmask(S,\omega))\,\big].
    \label{eq:coalition}
\end{equation}
With $v(\emptyset)$ defined, the Shapley/Owen axioms hold automatically, and attributions aggregate by additivity from features to regions and, for multivariate signals, from regions to channels (Section~\ref{sec:exp:covariates}). The codec is pluggable through three domain-blind interfaces; Table~\ref{tab:codecs} lists the ones we use.

\begin{table}[b]
\centering
\caption{\textbf{Codec zoo}: each is an interpretable encoder paired with a conditional generator. All are general except \emph{clinical} (ECG); \emph{shape} is the default.}
\label{tab:codecs}
\footnotesize
\begin{tabular}{@{}ll l@{}}
\toprule
Codec & Generator & Named features \\
\midrule
shape        & GP / cVAE   & level, trend, scale, peak \\
stat         & GP          & level, scale, slope \\
spectral-mix & GP          & per-band energy/width \\
clinical     & Mat\'ern GP & R-amp, QRS-width, ST/QT \\
\bottomrule
\end{tabular}
\end{table}

\paragraph{Intervention and reference semantics.} The interventional default draws absent features from the marginal $p_0$, isolating what the model uses, while the conditional draw takes them from $p(\theta_{\bar S}\mid\theta_S)$, crediting shared information. This is referred to as the interventional-versus-observational
distinction~\cite{janzing2020feature}. A marginal reference is assumption-free, whereas a class-conditional reference draws the absent value from a labeled cohort: the normal cohort (default), so each $\phi_i$ reads as its deviation from normal; the sample's own class; or a foil, giving the contrastive ``why this class rather than another.'' Conditional draws use a Gaussian copula that stays deterministic in $u$ (so CRN coupling survives) and collapses exactly to the marginal when the conditioning set is empty, so any difference is attributable to conditioning alone; App.~\ref{sec:refabl} compares the four references.


\subsection{Playing the Game On-manifold}\label{sec:play}

Playing the game means evaluating coalition values, each of which requires regenerating the absent regions on the data manifold. The codec's generator makes ``feature absent $\to$ regenerate region'' well-defined. 

Our default generator is a Gaussian process (GP) posterior: a distribution over plausible region shapes given the retained features and the region's boundary. A present feature conditions the GP on its value, an absent feature is set to the reference, and the region is sampled from that posterior. This keeps every coalition on the data manifold. For example, a spectrally and autocorrelation-matched GP sample of an ECG resembles the original form rather than a flat or noisy filler. The generator is swappable. Replacing it changes only the region sampler, not the game, estimator, or residual. A learned conditional VAE~\cite{kingma2014vae,sohn2015conditional} keeps the analytic encoder but replaces the GP sampler, improving off-manifold behavior at a small on-manifold cost, and swapping only the GP kernel (RBF, Mat\'ern, or spectral) alters the counterfactual's texture for robustness (App.~\ref{sec:kernels}). The exact GP game is our on-manifold reference: its closed-form posterior is what keeps an exact Owen sweep affordable, since a learned generator pays a per-coalition sampling cost it cannot amortize.

\paragraph{Owen values and exact completeness.} We attribute feature importance with Owen values, the average marginal contribution of a region-feature under permutations that preserve the block (region) structure. Since features are nested in regions, their Owen values roll up consistently. For instance, values of the QRS region's R-amplitude and width should sum to the QRS's Shapley value (Owen consistency), so the ``regions $\to$ features'' explanation is coherent. A telescoping common-random-number (CRN) sweep estimates every player from $|N|{+}1$ model calls along one nested chain:
\begin{equation}
\phi_i \;\mathrel{+}=\; f(\xmask(S\cup\{i\},\omega)) - f(\xmask(S,\omega)).
\label{eq:sweep}
\end{equation}
Reusing the same $\omega$ across adjacent coalitions, each term measures the effect of toggling feature $i$ rather than the gap between two noise draws. Antithetic pairing and a cached unit-variance Cholesky factor ($O(n^3)\!\to\!O(n^2)$) reduce cost further. The estimator is efficient at any budget: $\sum_i\phi_i=v(N)-v(\emptyset)$ holds to machine precision because it follows from the telescoping sweep, not from averaging permutations.

The reconstruction residual $\rho=f(x)-v(N)$ is the part of the decision that the feature representation cannot reproduce, since even the all-present coalition regenerates $x$ only approximately. We are required to always report $\rho$ because neglecting it breaks efficiency. Additionally, a large $\rho$ indicates an off-manifold generator. Therefore, to make the accounting exactly complete, we add a lossless morphology-residual player per region that restores the region's exact residual when present ($v(N)=f(x)$, $\rho\to0$) and regenerates normally when absent, so the opaque residual becomes one attributable named channel. Finally, to recover within-region resolution, we distribute each $\phi_r$ across its timesteps by a normalized occlusion profile ($\sum_{t\in r}\text{saliency}_t=\phi_r$); in Example~\ref{sec:intro}, this concentrates saliency at the R-peak within QRS.

\paragraph{Region discovery with \auto} When no region prior is available, the segmenter must find regions itself. The proposed segmenter model adopts three sources of information: a prior, when available; Pruned Exact Linear Time (PELT) change points~\cite{killick2012optimal}; and a model-aware segmenter. The signal-only default is PELT change, which reacts only to the signal. Our model-aware segmenter instead learns a selector (trained REAL-X-style~\cite{jethani2021have} with the generator as imputer) that reads regions from a thresholded mask. A raw selector might place boundaries within meaningful regions, where the codec regenerates from ambiguous fluctuations. We move each boundary at inference to the nearest clean-cut point, minimizing a cleanliness cost $c(t)$ (the distance to a PELT boundary or to an edge of the generator's informativeness profile), without retraining.

\subsection{Amortizing for Deployment with \amort{}}\label{sec:opt}

The exact GP game of Section~\ref{sec:play} is faithful but expensive: each coalition value invokes the generator and the model, and an exact Owen sweep evaluates many coalitions per instance. Because the game is a value function rather than a solver, its region-Owen attribution can be amortized. We train a single-forward-pass network $g_\psi$ against the game with the FastSHAP objective~\cite{jethani2021fastshap} and enforce the additive-efficiency constraint $\sum_i\phi_i=f(x)-v(\emptyset)$ by projection, so completeness holds exactly at inference rather than only in expectation. The resulting \amort{} runs $\sim$$10^3\times$ faster than the exact sweep while staying exactly complete.

The speedup comes at a cost in faithfulness. Because the surrogate is trained against a cross-sample removal operator, it is faithful to that operator: faithful off-manifold and, on real structured data, competitive with the exact game on-manifold. Its on-manifold faithfulness holds only where the signal carries sufficient structure for the surrogate to learn; on low-signal synthetic data it degrades, a limitation we report explicitly (App.~\ref{app:amortized}). We therefore recommend \amort{} for deployment on real data, and retain the exact GP game as our on-manifold reference throughout.
\section{Experimental Evaluation}\label{sec:exp}

We evaluate \method{} on six datasets across three domains: two synthetic families with ground-truth region masks (\textsc{Synth} and \textsc{Mixed}, which add static covariates); two clinical ECG families with cardiologist P/QRS/T annotations (\textsc{ECG}, from MIT-BIH~\cite{moody2001impact} and QT~\cite{laguna1997database}; and multivariate \textsc{LUDB}~\cite{kalinke2023ludb}, carrying patient age and sex; and two industrial-sensor families without region annotation (\textsc{FordA}, \textsc{Wafer}). All targets are InceptionTime~\cite{fawaz2020inceptiontime} classifiers.

We invoke \method{} in three configurations: \first provided regions; \second discovered regions with \auto{}; and \third amortized deployment with \amort{}, against six time series-specific windowing and perturbation baselines (WindowSHAP~\cite{nayebi2023windowshap}, TimeSHAP~\cite{bento2021timeshap}, TS-MULE~\cite{schlegel2021tsmule}, LIMESegment~\cite{sivill2022limesegment}, Dynamask~\cite{crabb2021dynamask}), the closest concept- and group-level Shapley players (C-SHAP~\cite{jutte2025cshap}, GS-SHAP~\cite{kim2026gsshap}), and non-time series gradient and sampling attributions (Integrated Gradients~\cite{sundararajan2017axiomatic}, gradient$\times$input~\cite{simonyan2014deep}, KernelSHAP~\cite{villani2022kernelshap}). 

Since perturbation faithfulness depends on how evidence is removed, we score each method under two matched operators: a neutral off-manifold operator and an on-manifold regeneration operator. We additionally report localization, robustness, cost, and completeness; every claim is gated on a paired Wilcoxon test with family-wide Benjamini--Hochberg $q<0.05$. Setup details are in Appendix~\ref{app:setup}.

\paragraph{In summary,} the exact game is the most faithful method on-manifold on every family and exactly complete (Section~\ref{sec:exp:faithful}); reporting both operators exposes the teaching-to-the-test trap, where the exact game still wins on real data under the comparators' own operator (Section~\ref{sec:exp:fair}); signal and static covariates are attributed on one common scale, so a model's reliance on metadata is read directly off the explanation (Section~\ref{sec:exp:covariates}). The method is robust to its design choices and discovers regions unaided (Section~\ref{sec:exp:robust}), while direct diagnostics show the generator's uncertainty is miscalibrated even where its attributions are most faithful (Section~\ref{sec:exp:generator}); and the amortized surrogate recovers all of this at lower cost on real data (Section~\ref{sec:exp:deploy}).

\begin{table*}[t]
\centering
\caption{\textbf{Method $\times$ metric across 6 families.} faith$_{\text{on}}$ $\uparrow$, faith$_{\text{off}}$ $\uparrow$. \best{Bold} = best in column, \runner{underline} = runner-up. Sample sizes: Synth $n{=}$30, Mixed $n{=}$30, ECG $n{=}$100, LuDB $n{=}$20, FordA $n{=}$24, Wafer $n{=}$24. Every method within a family is scored on the same instances from a single run. $^{\dagger}$ = \method{} beats every baseline for every dataset with statistical significance BH $q<0.05$. Paired two-sided Wilcoxon signed-rank test on per-instance differences. FordA and Wafer carry no region annotation.}
\label{tab:all_results}
\footnotesize
\begin{tabular}{@{}lcccccccccccc@{}}
\toprule
& \multicolumn{2}{c}{synth} & \multicolumn{2}{c}{mixed} & \multicolumn{2}{c}{ecg} & \multicolumn{2}{c}{ludb} & \multicolumn{2}{c}{forda} & \multicolumn{2}{c}{wafer} \\
\cmidrule(lr){2-3}\cmidrule(lr){4-5}\cmidrule(lr){6-7}\cmidrule(lr){8-9}\cmidrule(lr){10-11}\cmidrule(lr){12-13}
Method & faith$_{\text{on}}$ & faith$_{\text{off}}$ & faith$_{\text{on}}$ & faith$_{\text{off}}$ & faith$_{\text{on}}$ & faith$_{\text{off}}$ & faith$_{\text{on}}$ & faith$_{\text{off}}$ & faith$_{\text{on}}$ & faith$_{\text{off}}$ & faith$_{\text{on}}$ & faith$_{\text{off}}$ \\
\midrule
\method{} & \best{0.872}$^{\dagger}$ & -0.184 & \best{0.921}$^{\dagger}$ & 0.219 & \best{0.828}$^{\dagger}$ & 0.374 & \best{0.657}$^{\dagger}$ & 0.535 & \best{0.864}$^{\dagger}$ & 0.048 & \runner{0.542}$^{\dagger}$ & -0.079 \\
\auto{} & \runner{0.774} & -0.086 & \runner{0.850} & -0.051 & 0.496 & 0.191 & \runner{0.485} & 0.204 & \runner{0.600} & -0.046 & \best{0.774} & 0.102 \\
\midrule
windowshap (stat.) & -0.298 & 0.548 & 0.070 & 0.275 & 0.625 & \runner{0.506} & 0.284 & 0.525 & 0.089 & 0.386 & 0.020 & \runner{0.157} \\
windowshap (dyn.) & -0.268 & \runner{0.553} & 0.103 & 0.313 & 0.646 & \best{0.517} & 0.463 & \runner{0.574} & -0.019 & \runner{0.406} & 0.047 & \best{0.335} \\
timeshap & -0.303 & 0.550 & 0.566 & 0.316 & 0.591 & 0.440 & 0.437 & 0.561 & 0.025 & 0.386 & 0.039 & -0.022 \\
ts-mule (unif.) & -0.286 & 0.550 & 0.595 & 0.330 & 0.604 & 0.418 & 0.407 & 0.518 & 0.053 & \best{0.410} & 0.092 & 0.045 \\
ts-mule (sax) & -0.295 & 0.541 & 0.595 & 0.295 & 0.637 & 0.439 & 0.396 & 0.519 & 0.078 & 0.356 & 0.094 & -0.046 \\
limesegment & -0.290 & 0.541 & 0.576 & \runner{0.363} & 0.507 & 0.304 & 0.274 & 0.282 & 0.093 & 0.355 & 0.049 & 0.110 \\
dynamask & -0.114 & 0.083 & 0.091 & 0.009 & 0.616 & 0.366 & 0.208 & 0.128 & -0.051 & 0.231 & 0.069 & 0.037 \\
\midrule
c-shap & -0.397 & 0.484 & 0.401 & 0.217 & 0.659 & 0.402 & 0.437 & 0.295 & -0.023 & 0.231 & -0.000 & 0.048 \\
gs-shap & -0.361 & 0.525 & 0.646 & 0.267 & 0.453 & 0.335 & 0.168 & 0.286 & 0.060 & 0.133 & 0.042 & -0.095 \\
\midrule
integrated grad. & -0.288 & \best{0.568} & 0.526 & 0.358 & 0.518 & 0.426 & 0.317 & 0.406 & 0.077 & 0.311 & 0.007 & 0.081 \\
gradient$\times$input & -0.293 & 0.530 & 0.598 & 0.309 & 0.427 & 0.266 & 0.134 & 0.148 & 0.047 & 0.265 & 0.113 & -0.266 \\
kernelshap & -0.227 & 0.491 & 0.149 & \best{0.385} & \runner{0.662} & 0.478 & 0.462 & \best{0.611} & 0.043 & 0.405 & 0.090 & 0.152 \\
\bottomrule
\end{tabular}
\end{table*}

\subsection{Explanation Faithfulness and Completeness}\label{sec:exp:faithful}

On-manifold, where every method is scored under the same generative removal it is asked to explain, the exact game is the most faithful method on every family (Table~\ref{tab:all_results}); here faithfulness is the rank correlation between a player's attributed importance and the model's response when that player is removed by on-manifold regeneration~\cite{bhatt2020evaluating}. Provided regions (\method{}) leads on \textsc{Synth} ($0.872$), \textsc{Mixed} ($0.921$), \textsc{ECG} ($0.828$), \textsc{LuDB} ($0.657$), \textsc{FordA} ($0.864$), and discovered regions (\auto{}) leads on \textsc{Wafer} ($0.774$). For each dataset, \method{} outperforms all twelve baselines. The margin is significant against all twelve baselines jointly (Benjamini--Hochberg $q<0.05$ over the $144$ tests). This exact game is the reference standard against which the amortized surrogate is measured in 
Section~\ref{sec:exp:deploy}.

\begin{figure*}[t]
\centering
\includegraphics[width=0.8\textwidth]{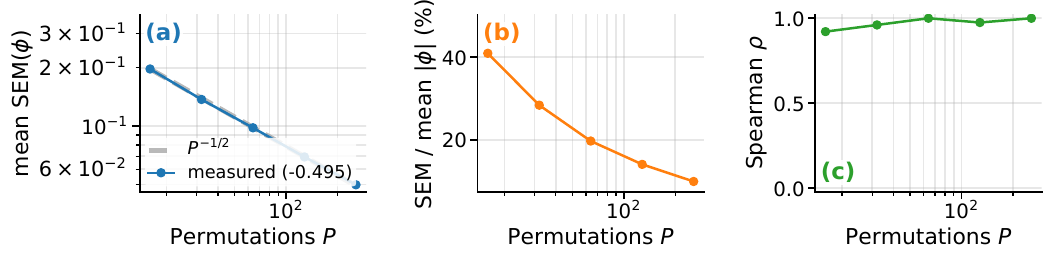}
\caption{\textbf{Owen estimator convergence in the permutation budget $P$}, on a fixed game
(\textbf{synth}; regions, codec and reference held fixed, only the estimator re-run).
\textbf{(a)} the standard error falls as $P^{-1/2}$ (fitted slope $-0.495$; the grey band is the
Monte-Carlo ideal, which the measured curve sits on top of).
\textbf{(b)} the same noise relative to mean $|\phi|$.
\textbf{(c)} region-ranking agreement with the largest-$P$ run. The budget is measured, not
assumed.}
\label{fig:converge}
\end{figure*}

Since in the exact game $\rho{=}0$, each attribution is an Owen value estimated by a common-random-number (CRN) sweep over a permutation budget $P$. Therefore, the estimate must be tight within the provided budget. Figure~\ref{fig:converge} depicts $P$ on a fixed \textsc{Synth} game. In this game, the per-player standard error falls at the Monte-Carlo-optimal rate. At the operating budget of $\bar P{=}128$, the region ranking already matches the largest-$P$ run at Spearman correlation $0.98$, with a noise floor of about $14\%$ of mean $|\phi|$. So ``exact'' names the estimand, the true two-level Shapley value rather than a windowing or gradient surrogate, and the residual is estimation variance that shrinks at the optimal rate with budget, not a modeling approximation.

Completeness is exact by construction. \method{} adds a residual player that closes the gap $\rho$ between the summed attributions and the model output, so $v(N){=}f(x)$ and the mean absolute residual $\overline{|\rho|}$ is $0.000$ on every family, for both the exact game and the amortized surrogate. This matches the additive KernelSHAP-family baselines. The surrogate inherits exact completeness from the efficiency projection rather than re-deriving it (Section~\ref{sec:opt}).

\begin{table}[t]
\centering
\caption{\textbf{Multivariate LUDB} (LVH task, $n{=}20$, $P{=}48$/$Q{=}2$; $\pm$: between-seed SEM). Three leads beat one on every quality metric in both regimes, at $\approx$$39\times$ runtime. Completeness stays exact. $\dagger$ localization is \emph{not} comparable across columns.}
\label{tab:mv}
\footnotesize
\begin{tabular}{@{}llcc@{}}
\toprule
 & & 3 leads (I/II/V1) & 1 lead (II) \\
\midrule
\multirow{5}{*}{\rotatebox{90}{provided}}
 & faith$_{\text{on}}$ $\uparrow$   & \best{0.735}\,\tiny{$\pm$.003} & 0.659\,\tiny{$\pm$.002} \\
 & faith$_{\text{off}}$ $\uparrow$  & \best{0.591}\,\tiny{$\pm$.028} & 0.370\,\tiny{$\pm$.006} \\
 & loc-AUPRC$^{\dagger}$            & 0.745\,\tiny{$\pm$.024} & 0.514\,\tiny{$\pm$.031} \\
 & $\overline{|\rho|}$              & $0.000$ & $0.000$ \\
 & runtime (s)                      & 94.4 & \best{2.4} \\
\midrule
\multirow{5}{*}{\rotatebox{90}{discov.}}
 & faith$_{\text{on}}$ $\uparrow$   & \best{0.511}\,\tiny{$\pm$.054} & 0.454\,\tiny{$\pm$.046} \\
 & faith$_{\text{off}}$ $\uparrow$  & \best{0.175}\,\tiny{$\pm$.039} & 0.154\,\tiny{$\pm$.055} \\
 & loc-AUPRC$^{\dagger}$            & 0.535\,\tiny{$\pm$.016} & 0.234\,\tiny{$\pm$.004} \\
 & $\overline{|\rho|}$              & $0.000$ & $0.000$ \\
 & runtime (s)                      & 260.8 & \best{6.9} \\
\bottomrule
\end{tabular}
\end{table}

\subsection{Comparison against Other Methods}\label{sec:exp:fair}


The perturbation and windowing baselines build players solely from time. Two recent methods build them differently and are the closest comparators to our region-and-covariate construction: C-SHAP, whose players are decomposition concepts (trend, bias, scale, variance, and low/high frequency) with no position, and GS-SHAP, whose players are feature-group $\times$ time-interval cells found by HSIC grouping and MMD segmentation. Neither releases code, so both are our reimplementations (Appendix~\ref{app:discussion}). Both remove a player by off-manifold mean or splice replacement, and both are exactly complete, so they are honest comparators rather than approximations we could beat on accounting alone.

On \textsc{Synth} the two operators point in opposite directions, and significantly so. The exact game beats both comparators on-manifold, and loses to both off-manifold. A method whose players are removed by mean or splice replacement is scored most favorably by a metric that removes evidence the same way; that is the teaching-to-the-test failure the two-operator protocol exists to expose, and it is why we report both columns for every method rather than choosing one.

On \textsc{ECG}, the disagreement disappears, and the exact game wins on-manifold, while preserving an approximate performance off-manifold. With real structure to model, \method{} leads on-manifold by $\approx0.17$ over C-SHAP, and $0.375$ over GS-SHAP. Meanwhile, the off-manifold method underperforms C-SHAP by $0.028$ and outperforms GS-SHAP by $0.039$. This can be interpreted as statistically insignificant. Therefore, on-manifold advantage is not an artifact of scoring the method under its own metric. The exact game performs well under baselines' preferred operator too.

\subsection{Temporal Signal and Covariates on One Scale}\label{sec:exp:covariates}

The cooperative game admits static covariates as singleton players (Section~\ref{sec:define}), so signal regions and metadata share a single scale, and a model's reliance on it is read off rather than inferred. On the LVH task, with sex and age as players at matched instances, region support, and budget ($n{=}40$), age receives mean $|\phi|=1.663$, above the top ECG region ($1.542$; QRS $1.430$), while sex, a near-distractor, receives $0.238$; together the statics carry $24.8\%$ of the total $|\phi|$, and age is the highest-ranked player on $62.5\%$ of instances. Admitting the statics costs no on-manifold faithfulness ($0.598$ to $0.597$, $p{=}0.89$) and leaves completeness exact, but it drops the expected-region hit rate from $0.575$ to $0.325$ ($p{=}0.002$): QRS loses top rank to age exactly where the model leans on age. Appendix~\ref{app:statics} gives the per-player table, a synthetic distractor control, and the full ablation.

The same construction extends to multichannel signals, in which the game places every lead on a single scale (Section~\ref{sec:define}). We run the LVH task on LUDB twice under an otherwise identical configuration, once on leads I/II/V1 and once on lead II alone, at the same reduced Owen budget ($P{=}48$, $Q{=}2$), the same $n{=}20$ instances and seeds, and both region regimes (Table~\ref{tab:mv}). Three leads beat one on every quality metric in both regimes (on-manifold faithfulness $0.659$ to $0.735$ with provided regions, $0.454$ to $0.511$ with discovered ones) at about $39\times$ the runtime ($2.4$ to $94.4$\,s). The player count roughly triples, from $64$ to $104$ effective players, and each coalition assembly regenerates three times as many regions, so a roughly quadratic blow-up is the expected shape of the cost rather than a surprise; the unaided arm pays worse than the exact one ($6.9$ to $260.8$\,s), because PELT re-segments every lead independently and produces more regions to price. 

The channel roll-up is exact by additivity, $\sum_c\sum_{i\in c}\phi_i$ plus the statics returns $\Sigma\phi$ without renormalization, and it recovers the clinically expected lead unprompted: averaged over instances the share of $|\phi|$ is V1 $0.398$, I $0.366$, II $0.235$, so V1 dominates and lead II is least informative, exactly what the Sokolow--Lyon criterion says, since it reads S in V1 and does not use lead II. This also flags that our other LUDB results, computed on lead II, use the least informative lead for this task. One localization number needs care: the multi-lead score is measured against the union of the expected region across leads with saliency summed over channels, so part of the $0.514$ to $0.745$ gain is pooling rather than better placement; the comparable statement is within-setting, where provided regions localize far better than discovered ones ($0.745$ vs $0.535$).

\subsection{Region Discovery and Ablations}\label{sec:exp:robust}

When regions must be found unaided, what matters is faithful attribution, not boundary recovery. Boundary-snapping (Section~\ref{sec:play}, Table~\ref{tab:segmenters}) delivers this with no retraining: the learnable model-aware selector matches or beats the signal-only PELT default on synth, mixed, and ludb, localizing better at far fewer players. The headline is ludb: the un-snapped selector collapses to $0.34$, worst of any arm, and snapping recovers $0.73$, past the incumbent, so placement, not count, was the problem. The strongest result is comparative: a ground-truth prior (the annotated P/QRS/T mask) further improves faithfulness and localization and dominates every unaided arm, so we recommend the annotated segmenter when a prior exists and the snapped, model-aware one when it does not.

Further, in the appendix, we ablate the kernel choices and their impact on the method's decisions. The kernel sets the counterfactual's texture: RBF is the most on-manifold faithful arm on ECG ($0.872$), while the rougher Mat\'ern kernels are the most robust (max-sensitivity $0.72$/$0.74$ against RBF's $0.91$); we default to RBF and report both directions (Appendix~\ref{sec:kernels}). The reference $p_0$ barely moves faithfulness (the five arms lie within one between-seed SEM on all datasets), so choosing one is a decision about what an explanation means, not a lever on quality (Appendix~\ref{sec:refabl}).

Conditional semantics is not a weak effect but an exactly zero one: it reproduces the marginal arm to machine precision ($\max|\Delta|\approx 10^{-15}$) on every region-level metric, even on real ECG where the codec features correlate three times as strongly as on synthetic. The reason is structural: Owen block values only focus on whole-block coalitions, where no absent feature has a retained sibling to condition on. Thus, feature correlation cannot change which regions matter. Within-region redistribution remains invariant. Faithfulness does not change for all six datasets, and only trades localization precision against recall (Appendix~\ref{sec:salabl}).

\begin{table}[b]
\centering
\caption{\textbf{Region discovery, unaided.} Downstream on-manifold faith$_{\text{on}}$ per family (reduced-budget \method{} re-run over each arm's regions); \best{bold} = best. \texttt{masknet} is a model-aware selector snapped onto clean cut points at inference (no retraining); \texttt{pelt-l2} is the signal-only default.}
\label{tab:segmenters}
\footnotesize
\begin{tabular}{@{}lcccc@{}}
\toprule
Segmenter & synth & mixed & ecg & ludb \\
\midrule
pelt-l2 \emph{(signal-only)} & 0.711 & 0.864 & \best{0.555} & 0.683 \\
masknet \emph{(learnable+snap)} & \best{0.721} & \best{0.908} & 0.518 & \best{0.729} \\
\bottomrule
\end{tabular}
\end{table}

\subsection{Generator Fit and Calibration}\label{sec:exp:generator}

The counterfactual generator is a swappable module: a GP posterior by default, or a cVAE or a DDPM in its place, and all three leave the accounting exactly complete. Expressiveness is not free, though. The more learned the generator, the more its regenerations resemble the counterfactuals a splice-based operator would produce, blurring the on-manifold/off-manifold distinction the two-operator protocol relies on, so a more powerful generator can raise measured faithfulness while weakening the very contrast that makes on-manifold faithfulness meaningful.

Because of this, we judge the generator by direct diagnostics rather than by the downstream faithfulness it produces. A probability-integral-transform check, whether held-out observations fall in their predicted quantiles at the right rate, gives Kolmogorov-Smirnov distances of $0.23$ to $0.40$ from uniform: the generator's predictive uncertainty is miscalibrated exactly on the families where its attributions are most faithful. Passing downstream faithfulness is therefore weaker evidence of a good generative model than it appears, which is why we report the fit directly (Appendix~\ref{sec:genresults}).

\subsection{Efficiency of Amortized Deployment}\label{sec:exp:deploy}

The exact game's main cost is runtime, and it is not intrinsic to the game. Amortizing the same value function into a learned explainer (Section~\ref{sec:opt}) replaces the per-instance Owen sweep with a single forward pass, a $10^3\times$ reduction from the exact game's $17.9$\,s per ECG beat, and the surrogate inherits exact completeness from the efficiency projection rather than re-deriving it ($\overline{|\rho|}{=}0.000$). It stays faithful where it matters: on
real data its on-manifold faithfulness tracks the exact game (ECG $0.556$ vs.\ $0.872$) and pays a cost only on low-signal synthetic families, where there is little on-manifold structure for a single-pass predictor to amortize (Appendix~\ref{app:amortized}, Tables~\ref{tab:paid} and~\ref{tab:trade}).

We measure the amortized arm against a representative four-baseline subset, one per paradigm (window-Shapley, LIME, realistic-LIME, and gradient-mask), at the deployment budget $n_{\text{amo}}$. The trade is deliberate, and it is the same one the fast comparators make with their own approximations (Section~\ref{sec:exp:fair}): a single-pass explainer at deployment cost that keeps the exact game's completeness and its faithfulness on the data with structure to explain.
\section{Conclusion}\label{sec:conclusion}

Meaningful time-series explanation begins with the choice of the right unit, the region, and the non-temporal covariates a model uses alongside it. Having fixed the players, an explanation is governed by the definition of a "baseline" region, defined by a counterfactual. Instead of choosing a fixed filler, the proposed method generates an absent region on the data manifold. Then it attributes the logit via Owen values estimated with common random numbers, and always reports the residual. Since all players share a single scale, the model's overreliance on static data rather than time series can be assessed. Experiments show that the exact game is the most faithful on-manifold on every family and exactly complete. On real ECG data, \method{} almost reaches the same off-manifold performance as other baselines. Finally, we attempt to ease the deployment process by introducing an amortized surrogate that reduces runtime on real data; a faithfulness-aware segmenter that enables region discovery without ground truth; and a game that supports multi-lead signals by recovering the clinically expected leads.

\paragraph{Limitations.} The most important question we want to address is whether these explanations are useful to domain experts. Therefore, our next step is conducting a user study. Since the amortized surrogate is trained on a single model, the model-randomization sanity check requires retraining the surrogate. Although the method achieves high performance on non-stationary data, we need to explore additional signals that a GP cannot represent. The exact generator saturates the logit, which the residual $\rho$ detects, and the swappable cVAE and DDPM generators mitigate this effect. Automated region discovery is improved but still unsolved. Further lessons that generalize beyond \method{}, an expert-rule agreement analysis, and the scope of our reimplementations, diffusion arm, and multivariate claims are developed in Appendix~\ref{app:discussion}.


\bibliography{references}

@inproceedings{crabb2021dynamask,
    title={Explaining Time Series Predictions with Dynamic Masks},
    author={Crabb{\'e}, Jonathan and Van Der Schaar, Mihaela},
    booktitle={Proceedings of the 38th International Conference on Machine Learning},
    pages={2166--2177},
    year={2021},
    editor={Meila, Marina and Zhang, Tong},
    volume={139},
    series={Proceedings of Machine Learning Research},
    month={18--24 Jul},
    publisher={PMLR},
    url={https://proceedings.mlr.press/v139/crabbe21a.html},
}

@misc{villani2022kernelshap,
    title={Feature Importance for Time Series Data: Improving KernelSHAP}, 
    author={Mattia Villani and Joshua Lockhart and Daniele Magazzeni},
    year={2022},
    eprint={2210.02176},
    archivePrefix={arXiv},
    primaryClass={cs.LG},
    url={https://arxiv.org/abs/2210.02176}, 
}

@inproceedings{sivill2022limesegment,
    title={LIMESegment: Meaningful, Realistic Time Series Explanations},
    author={Sivill, Torty and Flach, Peter},
    booktitle={Proceedings of The 25th International Conference on Artificial Intelligence and Statistics},
    pages={3418--3433},
    year={2022},
    editor={Camps-Valls, Gustau and Ruiz, Francisco J. R. and Valera, Isabel},
    volume={151},
    series={Proceedings of Machine Learning Research},
    month={28--30 Mar},
    publisher={PMLR},
    url={https://proceedings.mlr.press/v151/sivill22a.html},
}

@inproceedings{schlegel2021tsmule,
    author="Schlegel, Udo and Vo, Duy Lam and Keim, Daniel A. and Seebacher, Daniel",
    title="TS-MULE: Local Interpretable Model-Agnostic Explanations for Time Series Forecast Models",
    booktitle="Machine Learning and Principles and Practice of Knowledge Discovery in Databases",
    year="2021",
    publisher="Springer International Publishing",
    address="Cham",
    pages="5--14",
    isbn="978-3-030-93736-2"
}

@misc{raykar2023tsshap,
    title={TsSHAP: Robust model agnostic feature-based explainability for time series forecasting}, 
    author={Vikas C. Raykar and Arindam Jati and Sumanta Mukherjee and Nupur Aggarwal and Kanthi Sarpatwar and Giridhar Ganapavarapu and Roman Vaculin},
    year={2023},
    eprint={2303.12316},
    archivePrefix={arXiv},
    primaryClass={cs.LG},
    url={https://arxiv.org/abs/2303.12316}, 
}

@article{jutte2025cshap,
    title={C-SHAP for time series: An approach to high-level temporal explanations},
    author={Annemarie Jutte and Faizan Ahmed and Jeroen Linssen and Maurice van Keulen},
    journal={ArXiv},
    year={2025},
    volume={abs/2504.11159},
    url={https://api.semanticscholar.org/CorpusID:277786928}
}

@article{shapley1953value,
    title={A value for n-person games},
    author={Shapley, Lloyd S.},
    journal={Contributions to the Theory of Games},
    volume={2},
    number={28},
    pages={307--317},
    year={1953},
    publisher={Princeton University Press}
}

@inproceedings{lundberg2017unified,
    title={A unified approach to interpreting model predictions},
    author={Lundberg, Scott M. and Lee, Su-In},
    booktitle={Advances in Neural Information Processing Systems (NeurIPS)},
    pages={4765--4774},
    year={2017}
}

@article{strumbelj2014explaining,
  title={Explaining prediction models and individual predictions with feature contributions},
  author={{\v{S}}trumbelj, Erik and Kononenko, Igor},
  journal={Knowledge and Information Systems},
  volume={41},
  number={3},
  pages={647--665},
  year={2014},
  publisher={Springer}
}

@inproceedings{sundararajan2017axiomatic,
  title={Axiomatic attribution for deep networks},
  author={Sundararajan, Mukund and Taly, Ankur and Yan, Qiqi},
  booktitle={Proc. Int. Conf. Machine Learning (ICML)},
  pages={3319--3328},
  year={2017}
}

@inproceedings{petsiuk2018rise,
  title={{RISE}: Randomized input sampling for explanation of black-box models},
  author={Petsiuk, Vitali and Das, Abir and Saenko, Kate},
  booktitle={British Machine Vision Conference (BMVC)},
  year={2018}
}

@inproceedings{fong2017interpretable,
  title={Interpretable explanations of black boxes by meaningful perturbation},
  author={Fong, Ruth C. and Vedaldi, Andrea},
  booktitle={Proc. IEEE Int. Conf. Computer Vision (ICCV)},
  pages={3429--3437},
  year={2017}
}

@inproceedings{bento2021timeshap,
  title={{TimeSHAP}: Explaining recurrent models through sequence perturbations},
  author={Bento, Jo{\~a}o and Saleiro, Pedro and Cruz, Andr{\'e} F. and Figueiredo, M{\'a}rio A. T. and Bizarro, Pedro},
  booktitle={Proc. ACM SIGKDD Int. Conf. Knowledge Discovery and Data Mining (KDD)},
  pages={2565--2573},
  year={2021}
}

@article{nayebi2023windowshap,
  title={{WindowSHAP}: An efficient framework for explaining time-series classifiers based on {Shapley} values},
  author={Nayebi, Amin and Tipirneni, Sindhu and Reddy, Chandan K. and Foreman, Brandon and Subbian, Vignesh},
  journal={Journal of Biomedical Informatics},
  volume={144},
  pages={104438},
  year={2023},
  publisher={Elsevier}
}

@inproceedings{ismail2020benchmarking,
  title={Benchmarking deep learning interpretability in time series predictions},
  author={Ismail, Aya Abdelsalam and Gunady, Mohamed and Corrada Bravo, H{\'e}ctor and Feizi, Soheil},
  booktitle={Advances in Neural Information Processing Systems (NeurIPS)},
  pages={6441--6452},
  year={2020}
}

@article{theissler2022explainable,
  title={Explainable {AI} for time series classification: A review, taxonomy and research directions},
  author={Theissler, Andreas and Spinnato, Francesco and Schlegel, Udo and Guidotti, Riccardo},
  journal={IEEE Access},
  volume={10},
  pages={100700--100724},
  year={2022},
  publisher={IEEE}
}

@article{fawaz2020inceptiontime,
  title={{InceptionTime}: Finding {AlexNet} for time series classification},
  author={Ismail Fawaz, Hassan and Lucas, Benjamin and Forestier, Germain and Pelletier, Charlotte and Schmidt, Daniel F. and Weber, Jonathan and Webb, Geoffrey I. and Idoumghar, Lhassane and Muller, Pierre-Alain and Petitjean, Fran{\c{c}}ois},
  journal={Data Mining and Knowledge Discovery},
  volume={34},
  number={6},
  pages={1936--1962},
  year={2020},
  publisher={Springer}
}

@article{moody2001impact,
  title={The impact of the {MIT-BIH} arrhythmia database},
  author={Moody, George B. and Mark, Roger G.},
  journal={IEEE Engineering in Medicine and Biology Magazine},
  volume={20},
  number={3},
  pages={45--50},
  year={2001},
  publisher={IEEE}
}

@inproceedings{laguna1997database,
  title={A database for evaluation of algorithms for measurement of {QT} and other waveform intervals in the {ECG}},
  author={Laguna, Pablo and Mark, Roger G. and Goldberger, Ary and Moody, George B.},
  booktitle={Computers in Cardiology},
  pages={673--676},
  year={1997}
}

@article{goldberger2000physiobank,
  title={{PhysioBank}, {PhysioToolkit}, and {PhysioNet}: components of a new research resource for complex physiologic signals},
  author={Goldberger, Ary L. and Amaral, Luis A. N. and Glass, Leon and Hausdorff, Jeffrey M. and Ivanov, Plamen Ch. and Mark, Roger G. and Mietus, Joseph E. and Moody, George B. and Peng, Chung-Kang and Stanley, H. Eugene},
  journal={Circulation},
  volume={101},
  number={23},
  pages={e215--e220},
  year={2000},
  publisher={American Heart Association}
}

@inproceedings{janzing2020feature,
  title={Feature relevance quantification in explainable {AI}: A causal problem},
  author={Janzing, Dominik and Minorics, Lenon and Bl{\"o}baum, Patrick},
  booktitle={Proc. Int. Conf. Artificial Intelligence and Statistics (AISTATS)},
  pages={2907--2916},
  year={2020}
}

@inproceedings{sundararajan2020many,
  title={The many {Shapley} values for model explanation},
  author={Sundararajan, Mukund and Najmi, Amir},
  booktitle={Proc. Int. Conf. Machine Learning (ICML)},
  pages={9269--9278},
  year={2020}
}

@inproceedings{kumar2020problems,
  title={Problems with {Shapley}-value-based explanations as feature importance measures},
  author={Kumar, I. Elizabeth and Venkatasubramanian, Suresh and Scheidegger, Carlos and Friedler, Sorelle},
  booktitle={Proc. Int. Conf. Machine Learning (ICML)},
  pages={5491--5500},
  year={2020}
}

@inproceedings{hooker2019benchmark,
  title={A benchmark for interpretability methods in deep neural networks},
  author={Hooker, Sara and Erhan, Dumitru and Kindermans, Pieter-Jan and Kim, Been},
  booktitle={Advances in Neural Information Processing Systems (NeurIPS)},
  pages={9737--9748},
  year={2019}
}

@article{covert2021explaining,
  title={Explaining by removing: A unified framework for model explanation},
  author={Covert, Ian and Lundberg, Scott M. and Lee, Su-In},
  journal={Journal of Machine Learning Research},
  volume={22},
  number={209},
  pages={1--90},
  year={2021}
}

@article{owen1972multilinear,
  title={Multilinear extensions of games},
  author={Owen, Guillermo},
  journal={Management Science},
  volume={18},
  number={5-part-2},
  pages={64--79},
  year={1972},
  publisher={INFORMS}
}

@inproceedings{wilson2013gaussian,
  title={Gaussian process kernels for pattern discovery and extrapolation},
  author={Wilson, Andrew Gordon and Adams, Ryan Prescott},
  booktitle={International Conference on Machine Learning (ICML)},
  pages={1067--1075},
  year={2013}
}

@book{rasmussen2006gpml,
  title={Gaussian Processes for Machine Learning},
  author={Rasmussen, Carl Edward and Williams, Christopher K. I.},
  publisher={MIT Press},
  year={2006}
}

@inproceedings{deyoung2020eraser,
  title={{ERASER}: A benchmark to evaluate rationalized {NLP} models},
  author={DeYoung, Jay and Jain, Sarthak and Rajani, Nazneen Fatema and Lehman, Eric and Xiong, Caiming and Socher, Richard and Wallace, Byron C.},
  booktitle={Proceedings of the 58th Annual Meeting of the Association for Computational Linguistics (ACL)},
  pages={4443--4458},
  year={2020}
}

@inproceedings{bhatt2020evaluating,
  title={Evaluating and aggregating feature-based model explanations},
  author={Bhatt, Umang and Weller, Adrian and Moura, Jos{\'e} M. F.},
  booktitle={Proceedings of the 29th International Joint Conference on Artificial Intelligence (IJCAI)},
  pages={3016--3022},
  year={2020}
}

@inproceedings{adebayo2018sanity,
  title={Sanity checks for saliency maps},
  author={Adebayo, Julius and Gilmer, Justin and Muelly, Michael and Goodfellow, Ian and Hardt, Moritz and Kim, Been},
  booktitle={Advances in Neural Information Processing Systems (NeurIPS)},
  pages={9525--9536},
  year={2018}
}

@inproceedings{jethani2021have,
  title={Have we learned to explain? {How} interpretability methods can learn to encode predictions in their interpretations},
  author={Jethani, Neil and Sudarshan, Mukund and Aphinyanaphongs, Yindalon and Ranganath, Rajesh},
  booktitle={International Conference on Artificial Intelligence and Statistics (AISTATS)},
  pages={1459--1467},
  year={2021}
}

@article{killick2012optimal,
  title={Optimal detection of changepoints with a linear computational cost},
  author={Killick, Rebecca and Fearnhead, Paul and Eckley, Idris A.},
  journal={Journal of the American Statistical Association},
  volume={107},
  number={500},
  pages={1590--1598},
  year={2012},
  publisher={Taylor \& Francis}
}

@inproceedings{jethani2021fastshap,
  title={FastSHAP: Real-Time Shapley Value Estimation},
  author={Jethani, Neil and Sudarshan, Mukund and Covert, Ian and Lee, Su-In and Ranganath, Rajesh},
  booktitle={International Conference on Learning Representations (ICLR)},
  year={2022}
}

@inproceedings{kingma2014vae,
  title={Auto-Encoding Variational Bayes},
  author={Kingma, Diederik P. and Welling, Max},
  booktitle={International Conference on Learning Representations (ICLR)},
  year={2014}
}

@inproceedings{sohn2015conditional,
  title={Learning Structured Output Representation using Deep Conditional Generative Models},
  author={Sohn, Kihyuk and Lee, Honglak and Yan, Xinchen},
  booktitle={Advances in Neural Information Processing Systems (NeurIPS)},
  year={2015}
}

@inproceedings{chen2018learning,
  title={Learning to Explain: An Information-Theoretic Perspective on Model Interpretation},
  author={Chen, Jianbo and Song, Le and Wainwright, Martin J. and Jordan, Michael I.},
  booktitle={International Conference on Machine Learning (ICML)},
  year={2018}
}

@article{kim2026gsshap,
    title={GroupSegment-SHAP: Shapley Value Explanations with Group-Segment Players for Multivariate Time Series},
    author={Kim, Jinwoong and Park, Sangjin},
    journal={ArXiv},
    year={2026},
    volume={abs/2601.06114},
    url={https://arxiv.org/abs/2601.06114}
}

@inproceedings{lopezalcaraz2025causalconcept,
    title={Explaining Time Series Classification Predictions via Causal Attributions},
    author={Lopez Alcaraz, Juan Miguel and Strodthoff, Nils},
    booktitle={IEEE International Conference on Tools with Artificial Intelligence (ICTAI)},
    year={2025},
    note={arXiv:2405.15871; code at https://github.com/AI4HealthUOL/CausalConceptTS}
}

@article{gretton2012kernel,
    title={A Kernel Two-Sample Test},
    author={Gretton, Arthur and Borgwardt, Karsten M. and Rasch, Malte J. and Sch{\"o}lkopf, Bernhard and Smola, Alexander},
    journal={Journal of Machine Learning Research},
    volume={13},
    number={25},
    pages={723--773},
    year={2012}
}

@inproceedings{gretton2005hsic,
    title={Measuring Statistical Dependence with Hilbert-Schmidt Norms},
    author={Gretton, Arthur and Bousquet, Olivier and Smola, Alex and Sch{\"o}lkopf, Bernhard},
    booktitle={Algorithmic Learning Theory (ALT)},
    pages={63--77},
    year={2005}
}

@inproceedings{ho2020ddpm,
    title={Denoising Diffusion Probabilistic Models},
    author={Ho, Jonathan and Jain, Ajay and Abbeel, Pieter},
    booktitle={Advances in Neural Information Processing Systems (NeurIPS)},
    year={2020}
}

@book{rasmussen2006gp,
    title={Gaussian Processes for Machine Learning},
    author={Rasmussen, Carl Edward and Williams, Christopher K. I.},
    publisher={MIT Press},
    year={2006}
}

@article{kalinke2023ludb,
    title={Lobachevsky University Electrocardiography Database},
    author={Kalyakulina, Alena I. and Yusipov, Igor I. and Moskalenko, Viktor A. and Nikolskiy, Alexander V. and Kosonogov, Konstantin A. and Osipov, Grigory V. and Zolotykh, Nikolai Yu. and Ivanchenko, Mikhail V.},
    journal={PhysioNet},
    year={2021},
    doi={10.13026/eegm-h675}
}

@inproceedings{taufiq2023manifold,
  title={Manifold Restricted Interventional Shapley Values},
  author={Taufiq, Muhammad Faaiz and Bl{\"o}baum, Patrick and Minorics, Lenon},
  booktitle={Proceedings of the 26th International Conference on Artificial Intelligence and Statistics},
  pages={5079--5106},
  year={2023},
  volume={206},
  series={Proceedings of Machine Learning Research},
  publisher={PMLR},
  url={https://proceedings.mlr.press/v206/taufiq23a.html}
}

@inproceedings{simonyan2014deep,
  title={Deep inside convolutional networks: Visualising image classification models and saliency maps},
  author={Simonyan, Karen and Vedaldi, Andrea and Zisserman, Andrew},
  booktitle={Workshop at International Conference on Learning Representations (ICLR)},
  year={2014}
}

\newpage
\appendix
\section{Related Work}\label{app:related}

\paragraph{Shapley attribution and the removal operator.}
Shapley values, originally proposed by \citet{shapley1953value}, form the foundation of the modern attribution framework SHAP~\cite{lundberg2017unified}. The framework satisfies four basic axioms: efficiency, symmetry, dummy, and additivity. The main modeling decisions of Shapley attributions are the ``absence'' and sampling of features~\cite{strumbelj2014explaining}. Building on top of the framework, \citet{sundararajan2020many} and \citet{kumar2020problems} introduce attribution ambiguities; \citet{janzing2020feature} represent the observational-vs-interventional choice as causal; \citet{covert2021explaining} parameterize removal-based explanations with a masking operator. To adapt the removal mechanism for time series, we implement feature removal with a conditional generator to stay on-manifold. Further, we extend the value function from a traditional Shapley game to a two-level Owen value~\cite{owen1972multilinear} over nested features with corresponding time-series regions.

\paragraph{Time-series explainers.}
Time-series-specific explanation methods treat time steps as players to produce attributions~\cite{theissler2022explainable}. However, as Example~\ref{intro:ex} suggests, timestep saliency might not be sufficiently expressive. To test this, we compare \method{} against five representative methods that operate at, or can be aggregated to, the region level: \textbf{WindowSHAP}~\cite{nayebi2023windowshap} explains fixed-sized windows, \textbf{TimeSHAP}~\cite{bento2021timeshap} perturbs sequence events with pruning, \textbf{TS-MULE}~\cite{schlegel2021tsmule} runs LIME over segmented series, \textbf{LIMESegment}~\cite{sivill2022limesegment} builds realistic time-series perturbations, and \textbf{Dynamask}~\cite{crabb2021dynamask} learns a soft temporal perturbation mask. We leverage a region-based approach to learn generators that sample regions on the manifold.

\paragraph{Concept- and group-level players.}
Alternatively, other methods approach the saliency problem by defining players differently. C-SHAP~\cite{jutte2025cshap} decomposes a series into high-level concepts and computes exact Shapley values over them, trading localization for interpretability: its players are global properties and carry no position. For the multivariate methods, GroupSegment-SHAP~\cite{kim2026gsshap} builds players as (feature group $\times$ time interval) cells, grouping variables by Hilbert--Schmidt independence~\cite{gretton2005hsic} and segmenting time by maximum mean discrepancy~\cite{gretton2012kernel}. While these methods adopt a similar two-level construction, they redefine the player and keep an off-manifold removal operator (mean replacement). Oppositely, we hold the player structure fixed and change the operator. Our harness scores both axes at once, so Table~\ref{tab:all_results} can separate them.

\paragraph{Hierarchical credit assignment.}
Nesting players is exactly what Owen values~\cite{owen1972multilinear} formalizes: a coalition structure partitions players into blocks, credit is allocated between blocks and then within them, and the result reduces to Shapley when the partition is trivial. We adopt that structure directly---features nested in regions, and regions nested in channels for multivariate signals---rather than building a bespoke coarse-to-fine procedure, which buys the consistency and efficiency guarantees for free.

\paragraph{Generative imputation for counterfactual concepts.}
We build on the notion that counterfactuals are generated rather than spliced. Thus, our generator operates only on a single region conditioned on its retained features and boundary, allowing us to keep exact Owen value computation affordable. We include a region-scale conditional DDPM~\cite{ho2020ddpm} as a third generator family (Section~\ref{app:impl}) to test whether the choice of generator, rather than the decision to generate at all, is what matters.

\paragraph{Generators and backbones.}
Our default generator is a Gaussian process~\cite{rasmussen2006gp} with a spectral-mixture kernel~\cite{wilson2013gaussian} and a beat-template mean for ECG; a learned conditional VAE~\cite{kingma2014vae,sohn2015conditional} and a conditional DDPM~\cite{ho2020ddpm} are the alternatives we explored behind the same game. Segmentation change points are identified via PELT~\cite{killick2012optimal}. As a predictive baseline, we leverage InceptionTime~\cite{fawaz2020inceptiontime}.

\paragraph{Amortized and learned explanation.}
Solving an exact game for every instance is computationally expensive, which motivated training learnable networks to predict explanations. FastSHAP~\cite{jethani2021fastshap} amortizes Shapley values with an efficiency-constrained regression loss, L2X~\cite{chen2018learning} and REAL-X~\cite{jethani2021have} learn instance-wise feature selectors. We leverage the FastSHAP objective while modifying the model to account for time-series specifics. This allows the deployment of the expensive exact method.

\paragraph{On- vs off-manifold evaluation.}
Perturbation-based faithfulness is defined by corresponding methods' feature removal mechanisms~\cite{petsiuk2018rise,fong2017interpretable}. Scoring an explainer only with an off-manifold operator rewards methods that measure importance in that same space, \ie the ``teaching-to-the-test'' trap~\cite{hooker2019benchmark,jethani2021have}. We report faithfulness under \emph{both} a neutral off-manifold operator and a matched on-manifold one, applied identically to every method.

\section{Implementation details}\label{app:impl}
In the following section, we define the default implementation details for the experiments presented.

\paragraph{Codec feature encoders.}
Aiming to produce encodings for the generators, we propose codecs. A codec is a pair of an encoder $\phi$ and a conditional generator. The encoder is a closed-form read of the region without fitting and learned parameters. Thus, we can maintain the same feature meaning across all regions and signals. \texttt{stat} encodes \texttt{level} (region mean), \texttt{slope} (least-squares linear coefficient in $t$), and \texttt{scale} (standard deviation of the residual after removing level and slope). \texttt{shape} is a strict superset that adds a named extremum: \texttt{peak\_amp}, the signed height of the dominant extremum above the fitted line, and \texttt{peak\_loc}, its position as a fraction of the region span in $[0,1]$ normalized in a way that regions of different lengths pool into one reference distribution. \texttt{shape}'s \texttt{scale} is calibrated on the residual after the fitted peak, so the generator's variance matches what the named features leave behind. \texttt{basis} encodes low-order DCT coefficients; \texttt{sm} encodes spectral-mixture hyperparameters; \texttt{nonstationary} encodes a per-region lengthscale profile. Under \texttt{GRAIL\_stabilized}, the slope is a Theil--Sen estimate instead of least squares. The generator's mean is the encoder read backwards ($\text{level} + \text{slope}\cdot(t-\bar t)$, plus $\text{peak\_amp}\cdot g(t)$ for \texttt{shape}, with $g$ a Gaussian bump of width $\max(1.5, 0.08L)$ at $\text{peak\_loc}\cdot(L{-}1)$); the residual is a boundary-conditioned GP draw. All experiments here use \texttt{shape} with an RBF residual kernel.

\paragraph{Gaussian Process.}
The residual kernel has an amplitude, fixed by the encoded \emph{scale}, and a lengthscale $\ell = f\cdot L$ for region length $L$. The default $f = 0.25$ is a heuristic. Under \texttt{GRAIL\_empirical\_bayes}, it is instead selected by maximizing the mean per-sample marginal log-density over the grid $f\in\{0.03,0.06,0.10,0.15,0.25,0.35,0.50\}$, evaluated on the \emph{reference corpus} and cached per $(\text{codec}, L)$. Selection never sees the explained signal, so the value function stays independent of the instance. A $\sigma_n^2 = 10^{-2}\,\mathrm{var}(x_r)$ is added before the Cholesky. Given this value is not provided, the criterion ranks candidates by conditioning on fit (\S\ref{sec:play}). Jitter is $10^{-6}$, and the boundary flank width is $8$ samples on each side.

\paragraph{Reference corpus.}
$p_0$ is fitted once per run, before the attributions are provided. Then, with the segmenter preserved independently of a data split, regions are encoded. Resulting feature pools are based on the empirical marginals. A feature value is sampled by inverse-CDF from the corresponding pool. To preserve the on-manifold assumption, we restrict the pool conditionally by \texttt{class\_conditional}, label (normal cohort, target class, or foil-class, \S\ref{sec:refabl}). \texttt{Conditional} additionally fits a Gaussian copula and draws the absent coordinate from the Gaussian conditional given the retained features, evaluated at the same uniform $u$, such that CRN coupling survives. When neither retained features nor a single correlation block is kept, the algorithm falls back to the marginal, so \texttt{conditional} reduces to \texttt{marginal} exactly.

\paragraph{Estimator budget.}
The budget hyperparameters used in the experiments are as follows: restricted permutations $P{=}128$, base-randomness draws $Q{=}3$, exact enumeration whenever a block has $\le 10$ players, batch size $2048$, antithetic pairing on. To enhance the processing speed, we employ batched forward passes on coalitions. Figure~\ref{fig:converge} measures the resulting error: the mean per-player standard error (SME) falls as $P^{-0.495}$ against the Monte-Carlo ideal of $P^{-1/2}$, \ie CRN reduces the constant without changing the rate, and $P{=}128$ is located at the curve's flat span relative to its own cost. On the other hand, the DDPM generator runs at $P{=}16$, $Q{=}1$ (\S\ref{sec:play}) and the multivariate arm at $P{=}48$, $Q{=}2$.

\paragraph{Neural generators.}
The cVAE is a $64$-point grid autoencoder, $d_z{=}8$, hidden width $128$, trained $200$ epochs at $\mathrm{lr}{=}10^{-3}$ with $\beta_{\text{KL}}{=}0.05$ and a cycle-consistency term of weight
$0.2$, on region crops from at most $400$ training signals. The DDPM shares the grid and the conditioning vector $\theta$, both boundary flanks, two presence flags, normalized length, optionally a class index, with $200$ training steps and a $20$-step strided DDIM reverse chain at sampling time. Both expose the identical $\texttt{sample}(\theta, \text{left}, \text{right}, L, \varepsilon)$ signature as the GP, and both are deterministic in $\varepsilon$. The reverse chain consumes slices of the same $\varepsilon$ as the CRN sweep, so toggling a feature measures the toggle, not a fresh noise draw. Regions are resampled to the grid and back by linear interpolation. \amort{} is a FastSHAP-style head trained on $200$ signals $\times$ $24$ coalitions. We ablate the operator that generates its targets in \S\ref{sec:distill}.

\section{A worked clinical example}\label{app:qualitative}
\begin{figure}[t]
\centering
\includegraphics[width=\columnwidth]{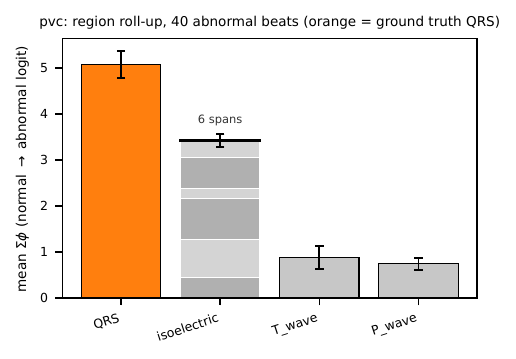}
\caption{\textbf{\method{} region roll-ups over $40$ abnormal PVC beats} (shape codec, RBF, deviation-from-normal reference; $\rho{=}0$ on every beat). Bars are the mean summed Owen value per region, $\pm$SEM; orange marks the clinical ground truth (QRS), which leads on $31/40$ beats. The isoelectric bar is the aggregation of background regions. Therefore, it is split into its six constituent spans.}
\label{fig:rollup}
\end{figure}

In Fig.~\ref{fig:regions}, $\phi$ is one (region, feature) player and $\Sigma\phi$ a region roll-up. Both are at different levels of the game and are never compared directly. Fig.~\ref{fig:rollup} shows the roll-up over $40$ abnormal beats of the PVC task under the default arm (shape codec, RBF, deviation-from-normal reference). The QRS complex, which is the clinical ground truth, carries the largest mean roll-up ($\Sigma\phi{=}{+}5.07{\pm}0.30$), leads on $31/40$ beats, and is positive on all $40$ beats. The aggregated isoelectric segment is second ($+3.42{\pm}0.14$), but only because it is the sum of six separate background spans ranging from $+0.21$ to $+0.89$. The largest is still almost 6 times below QRS, which is why Fig.~\ref{fig:rollup} splits that bar into the spans composing it. Based on the received attributions, we can algorithmically construct the explanation narrative as follows: ``\emph{the strongest driver toward class~1 is the QRS complex's elevated baseline/ST level (extremely, $4.9\sigma$; $\phi{=}{+}4.16{\pm}0.09$ log-odds); QRS contributes the most in total ($\Sigma\phi{=}{+}5.70$); $0\%$ of the decision is unexplained ($\rho{=}{+}0.0000$); within-signal saliency peaks at $t{=}110$, inside the QRS.}'' The PVC task is primarily determined by abnormalities in QRS morphology that highlight the importance of the most attributed region, thus proving the faithfulness of the algorithm. Completeness here is measured to be $\max|\rho|{<}10^{-4}$ across all $40$ beats.

\section{Kernel functions}\label{app:kernelx}\label{sec:kernels}

The generator module can be selected depending on the nature of the data. The kernel defines the counterfactual's underlying function and thus determines what a coalition looks like without pinned features. Our default RBF is infinitely smooth, yet it is expected to perform poorly for non-smooth signals; it showed the best performance in a non-stationary setting. Behind the identical game, we therefore compare three cheap variance-separable kernels: RBF, Mat\'ern-3/2, Mat\'ern-5/2~\cite{rasmussen2006gpml}, whose rougher sample paths track real signals while keeping the cached-Cholesky speed-up (\S\ref{sec:define})---and a \emph{filterbank} spectral-mixture codec that fixes a component's frequency to a log-spaced band grid, so its identity is stable under input jitter (at the cost of the non-separable solve). Every arm is exactly complete ($\overline{|\rho|}{=}0$).

\begin{figure*}[t]
\centering
\includegraphics[width=\textwidth]{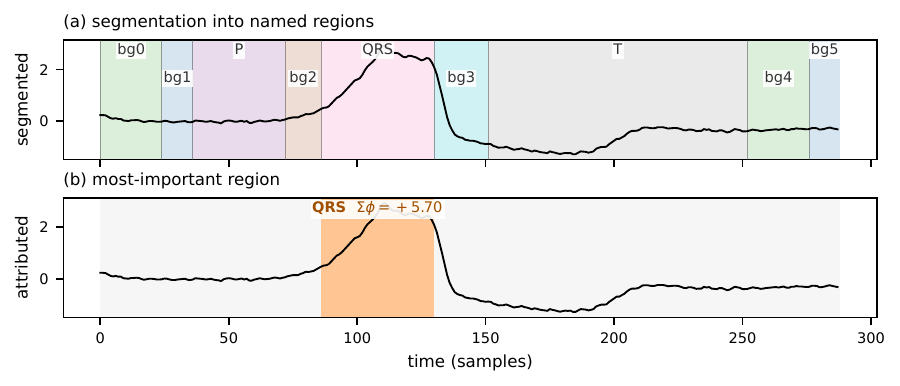}
\caption{\textbf{From signal to attributed regions} (\method{}, RBF, a PVC beat). (a) the beat's segmentation into named P/QRS/T waves plus the chunked isoelectric background (\texttt{bg0}--\texttt{bg5}); (b) the same beat with the top region by Owen value shaded. The QRS complex, which is the clinical ground truth for a PVC, is the most important region ($\Sigma\phi{=}{+}5.70$, a region roll-up), and the accounting is exactly complete ($\rho{=}{+}0.0000$). The beat is the median one among those with QRS leads in the Fig.~\ref{fig:rollup} cohort, where the typical case rather than the most favorable one holds, and both figures come from a single run of a single config.}
\label{fig:regions}
\end{figure*}

On ECG (Table~\ref{tab:kernels}), all four exact-game kernels are more on-manifold faithful than every baseline ($0.72$--$0.84$ vs.\ $\le 0.66$) and exactly complete. However, the trade-off for faithfulness is runtime speed, which may be one to three orders of magnitude slower. Localization is saliency-dependent: at this sweep's setting, the baselines edge ahead (timeshap loc-AUPRC $0.554$ vs.\ our best $0.519$), but under the shipped occ-$w{=}6$ default \method{} takes the ECG localization lead (Table~\ref{tab:all_results}, \S\ref{sec:salabl}). Among kernels, Mat\'ern-5/2 is the most faithful, and the rougher Mat\'ern kernels are the most robust, reaching a lower max-sensitivity than the smooth RBF ($0.72$/$0.74$ vs.\ $0.91$). Fig.~\ref{fig:regions} traces one PVC beat under the Mat\'ern-5/2 arm: segmentation into named cardiac regions, and the QRS complex correctly surfaced as the top region---an exactly-complete, clinically-right explanation. App.~\ref{app:kernelx} extends the ablation to non-stationary FordA and the mass-accounting mechanism. We default to RBF since on-manifold faithfulness is the primary objective in this work. The filterbank codec can be used primarily for jitter-stable component identity.

\begin{table}[ht]
\centering
\caption{\textbf{Kernel ablation vs.\ baselines on real ECG.} Exact-game kernels ($n{=}120$, ablation sweep) against the eight baselines ($n{=}200$, frozen full run): real numbers, reported side by side but not sample-paired (different $n$). On-manifold faithfulness $\uparrow$, localization loc-AUPRC $\uparrow$, max-sensitivity $\downarrow$, runtime~s $\downarrow$; every exact-game arm has $\overline{|\rho|}{=}0$ by construction. \best{Bold} = best in column. The exact game leads to faithfulness and completeness; the rougher Mat\'ern kernels are the most robust (lowest max-sensitivity).}
\label{tab:kernels}
\setlength{\tabcolsep}{4pt}
\footnotesize
\begin{tabular}{@{}lcccc@{}}
\toprule
Method & faith$_{\text{on}}$ & loc & m-sens & s \\
\midrule
\method{} (rbf)          & 0.831 & 0.480 & 0.908 & 12.2 \\
\method{} (mat\'ern-3/2) & 0.828 & 0.519 & \best{0.719} & 12.1 \\
\method{} (mat\'ern-5/2) & \best{0.841} & 0.514 & 0.744 & 12.2 \\
\method{} (filterbank)   & 0.723 & 0.445 & 0.767 & 40.0 \\
\midrule
windowshap (stat.)  & 0.634 & 0.542 & -- & 1.4 \\
windowshap (dyn.)   & 0.658 & 0.508 & -- & 2.7 \\
timeshap            & 0.624 & \best{0.554} & -- & 2.7 \\
ts-mule (unif.)     & 0.620 & 0.460 & -- & 0.02 \\
ts-mule (sax)       & 0.649 & 0.497 & -- & \best{0.02} \\
limesegment         & 0.591 & 0.468 & -- & 0.50 \\
tsshap-surrogate    & 0.469 & 0.447 & -- & 0.03 \\
dynamask            & 0.633 & 0.511 & -- & 1.5 \\
\bottomrule
\end{tabular}
\end{table}

\paragraph{Non-stationary data (FordA).}
On a non-stationary family (FordA engine-sensor, UCR; no region ground truth, so localization is undefined), the tradeoff is more evident. While the expressive kernels help achieve input robustness, i.e., max-sensitivity falls from RBF's $0.39$ to $0.21$ for the filterbank codec, with completeness still exact. However, the expressiveness entails lower on-manifold faithfulness. Particularly, RBF is the most faithful ($0.864$), while Mat\'ern-5/2 drops to $0.790$. The results here are the exact opposite of ECG, since FordA's discriminative signal is not preserved by the residual.

\paragraph{Importance of kernel selection.}
Since RBF is over-smoothing, the residual becomes large, preventing recovery of the target signal. On FordA, only $18\%$ of the attributed mass sits on named, interpretable features under RBF, versus roughly half under Mat\'ern-5/2, while the fixed-basis and spectral codecs push most of the mass onto named players. Matching the generator's roughness to the signal keeps the explanation interpretable, which is why Mat\'ern improves robustness in such settings. Therefore, we recommend \textbf{Mat\'ern-5/2} as the default for non-stationary settings. Meanwhile, RBF can be selected when the last of on-manifold faithfulness matters most. Finally, the filterbank codec can be used when stability under jitter is the priority.

\section{Experimental Setup}\label{app:setup}

\paragraph{Datasets.}
We use six datasets across three domains; region ground truth is used only to score localization and is never seen at explanation time except by the provided-regions arm. We define data sizes in the following manner: number of samples $\times$ number of channels $\times$ sequence size.

\textbf{Synth} is a synthetic family of $\TT{=}100$-step univariate series: i.i.d.\ background with localized injected features whose combinations define the target classes, and the injection masks are the ground-truth region attributions. Data size: $1500{\times}1{\times}100$.

\textbf{Mixed} extends \textbf{Synth} with static features acting as shortcut, informative, or distractor variables, giving a controlled setting with known signal \emph{and} known covariate ground truth. Data size: $1500{\times}1{\times}100$.

\textbf{ECG} uses the MIT-BIH Arrhythmia~\cite{moody2001impact} and QT~\cite{laguna1997database} databases from PhysioNet~\cite{goldberger2000physiobank}; each R-aligned beat carries cardiologist P/QRS/T regions across five binary tasks. Data size: $4000{\times}1{\times}288$.

\textbf{LUDB} is the Lobachevsky University Database~\cite{kalinke2023ludb}: real ECG with cardiologist P/QRS/T boundaries and per-subject age and sex, on the left-ventricular-hypertrophy (LVH) task; it is multivariate (leads I, II, V1). Data size: $1799{\times}3{\times}400$.

\textbf{FordA} is a UCR engine-sensor series, non-stationary and without region annotation (localization undefined). Data size: $4921{\times}1{\times}100$.

\textbf{Wafer} is a UCR semiconductor-sensor series, also without region
annotation. Data size: $7164{\times}1{\times}100$.

\paragraph{Target models.}
All explanations target InceptionTime~\cite{fawaz2020inceptiontime} classifiers trained per dataset. We train the model with Adam at a peak learning rate of $10^{-3}$ and a batch size of $64$. For the synth, ecg, forda, and wafer families, utilize the \texttt{fastai} one-cycle schedule (\texttt{fit\_one\_cycle}, $\text{lr}_{\max}{=}10^{-3}$) with a validation batch size of $128$ and a \texttt{TSStandardize} batch transform. The static-fusion families (mixed, ludb, and the 3-lead ludb build) use a plain \texttt{torch} loop at a constant $10^{-3}$ with no schedule. Every family minimizes cross-entropy. Epoch numbers are $50$ on synth and mixed, $30$ on ecg, $40$ on forda and wafer, and $60$ on both ludb builds.

\paragraph{Methods and configurations.}
\method{} uses the \emph{shape} codec on general families and the \emph{clinical} codec on ECG, a class-conditional (normal-cohort) reference, and the lossless morphology-residual player. It appears in three configurations: with a provided region prior (\method{}); discovering its own regions via the model-aware segmenter with boundary snapping (\auto{}, \S\ref{sec:play}); and as the amortized single-forward-pass deployment (\amort{}, \S\ref{sec:opt}). The perturbation and windowing baselines are WindowSHAP (stationary, dynamic)~\cite{nayebi2023windowshap}, TimeSHAP~\cite{bento2021timeshap}, TS-MULE (uniform, SAX)~\cite{schlegel2021tsmule}, LIMESegment~\cite{sivill2022limesegment}, TsSHAP~\cite{raykar2023tsshap}, and Dynamask~\cite{crabb2021dynamask}. C-SHAP~\cite{jutte2025cshap} and GS-SHAP~\cite{kim2026gsshap} do not release code. We have reimplemented them and made the code available in the supplementary materials. Gradient baselines are integrated gradients~\cite{sundararajan2017axiomatic}, gradient$\times$input, and KernelSHAP~\cite{lundberg2017unified}. The full head-to-head uses all configurations. The amortized-arm tables use a representative four-baseline subset (one per paradigm) at the reduced sample size $n_{\text{amo}}$. The baselines share the same region support and harness; \auto{} discovers its own.

\paragraph{Metrics.}
\emph{Faithfulness} is scored under two matched removal operators, a neutral \emph{off-manifold} splice and an \emph{on-manifold} regeneration, via comprehensiveness/sufficiency~\cite{deyoung2020eraser}, faithfulness correlation~\cite{bhatt2020evaluating}, and deletion/insertion AUC; the on- and off-manifold variants are reported side by side (the \texttt{\_on}/\texttt{\_off} columns). \emph{Localization} compares the per-timestep saliency to the annotated region mask~\cite{ismail2020benchmarking} by AUPRC and a pointing game. \emph{Robustness} covers shift/jitter stability, max-sensitivity, and the model-randomization sanity check~\cite{adebayo2018sanity}. \emph{Cost} is the effective-player count and wall-clock runtime; \emph{completeness} is $\overline{|\rho|}$; and \emph{self-segmentation} is boundary $F_1$ against the mask.

\paragraph{Statistical protocol.}
Within a family, every method is scored on the same instances from a single run, so all comparisons are genuinely paired. Every ``wins/loses'' claim is gated on a paired two-sided Wilcoxon signed-rank test on per-instance differences, corrected family-wide with Benjamini--Hochberg at $q<0.05$, and reported with the effect size. The sample sizes are provided in the description of Table~\ref{tab:all_results}.

\paragraph{Implementation.}
Codec feature encoders, GP kernels and hyperparameters (including the residual lengthscale, fitted by marginal likelihood over the reference corpus rather than the explained signal, and the observation-noise nugget $\sigma_n^2=10^{-2}\,\mathrm{var}$), reference-corpus construction and the Gaussian-copula conditional reference, the estimator budget ($P$, $Q$), and the cVAE and DDPM generator architectures and training are specified in Appendix~\ref{app:impl}.

\section{``Absent'' features}\label{sec:refabl}
The reference $p_0$ is one of the two switches (conditional, interventional) that change what an explanation means. Table~\ref{tab:reference} sweeps all five on synth at $n{=}80$ (20 instances $\times$ 2 seeds $\times$ 2 runs) and repeats the sweep on both real families ($n{=}80$ ecg, $n{=}16$ ludb), each in both region regimes.

\begin{table}[t]
\centering
\caption{Reference ablation, \texttt{faithfulness\_corr\_on} ($\pm$ between-seed SEM; $n{=}80$ synth, $80$ ecg, $16$ ludb). ``Provided'' uses ground-truth region support, ``discovered'' unaided segmentation.}
\label{tab:reference}
\footnotesize
\begin{tabular}{@{}lcccccc@{}}
\toprule
 & \multicolumn{2}{c}{synth} & \multicolumn{2}{c}{ecg} & \multicolumn{2}{c}{ludb} \\
\cmidrule(lr){2-3}\cmidrule(lr){4-5}\cmidrule(lr){6-7}
reference $p_0$ & prov. & disc. & prov. & disc. & prov. & disc. \\
\midrule
marginal      & 0.636 & 0.540 & 0.684 & 0.466 & 0.613 & 0.473 \\
normal cohort & 0.620 & \best{0.582} & 0.694 & 0.462 & \best{0.628} & 0.461 \\
target class  & 0.626 & 0.560 & \best{0.696} & \best{0.478} & 0.607 & \best{0.476} \\
foil class    & \best{0.638} & 0.581 & 0.669 & 0.466 & 0.617 & 0.468 \\
conditional   & 0.636 & 0.533 & 0.684 & 0.466 & 0.613 & 0.473 \\
\midrule
spread        & 0.018 & 0.049 & 0.027 & 0.016 & 0.021 & 0.015 \\
\bottomrule
\end{tabular}
\end{table}

\textbf{Importance of reference.} With ground-truth regions, the
five synth arms lie within roughly one between-seed SEM ($0.620$--$0.638$); the spread widens only under unaided segmentation ($0.533$--$0.582$), where the normal-cohort default leads. Real data agrees and is tighter still: on ecg the arms span $0.669$--$0.696$ (provided) and $0.462$--$0.478$ (discovered), on ludb $0.607$--$0.628$ and $0.461$--$0.476$. The choice of $p_0$ is therefore a choice about \emph{what the explanation means}---deviation from normal, from the predicted class, from the foil---and not a lever on measured faithfulness. That is a cleaner statement than ``our default is best'', and it is the one the data supports.

\textbf{Conditional semantics is zero.} The conditional arm reproduces the marginal arm on \emph{every} region-level metric, on all three families, to machine precision ($\max|\Delta|\approx10^{-15}$). Our first explanation was that the codec features are nearly independent, so a Gaussian conditional \emph{is} the marginal; that explanation is wrong, and real data refutes it. Copula correlation among the codec features is a property of the reference corpus that can be measured without running the ablation at all, and on real ECG it is about three times synth's: $\max|\rho_{ij}|$ of $0.41$--$0.49$ (mean $0.12$--$0.18$) across the five ecg tasks and $0.41$ (mean $0.14$) on ludb, against synth's $0.155$ (mean $0.070$). We predicted the conditional arm would separate there. It does not, by a margin of $10^{-15}$.

The real reason is a structural property of Owen values rather than of the data. Instrumenting the draw confirms that the conditional branch fires on $60\%$ of absent draws and the \emph{feature}-level Owen values do move, by up to $3.3\%$ of $\Sigma|\phi|$. What does not move is the \emph{region} roll-up. Owen block values telescope to the Shapley value of the game played over whole blocks, and in a whole-block coalition every region is either entirely retained or entirely absent, so an absent feature never has a retained sibling \emph{in its own region} to condition on. The conditional reference falls back to the marginal on precisely the coalitions the region roll-up is built from, so $\sum_{i\in r}\phi_i$ is identical by construction, at any feature correlation. Every metric in this paper consumes that roll-up, or a saliency profile derived from it, and is therefore blind to the switch by design rather than by accident.

This is worth stating plainly because it binds a whole class of proposed improvements. Conditional semantics is a within-region refinement. It changes how a region's credit is divided among its named features. It cannot change which regions are important. The class-conditional cohorts, by contrast, redraw the pool a fully-absent region is filled from, so they do change the block-level game, which is why
those three arms separate while this one cannot.

\section{Where a region's credit goes}\label{sec:salabl}
Within-region redistribution (\S\ref{app:impl}) preserves the region roll-up by construction, so it
can only move \emph{localization}---never faithfulness. Table~\ref{tab:saliency} confirms this
holds exactly in practice and shows what the choice does buy.

\begin{table}[t]
\centering
\caption{Within-region redistribution on \textbf{synth} ($n{=}80$). All six profile arms leave faithfulness bit-identical (0.620 provided / 0.582 discovered), as the roll-up invariant requires; only rank-alignment perturbs it, downward. ``occ-$w$'' is occlusion at window width $w$; ``+f'' adds the 0.25 uniform floor.}
\label{tab:saliency}
\footnotesize
\begin{tabular}{@{}lcccc@{}}
\toprule
 & \multicolumn{2}{c}{provided regions} & \multicolumn{2}{c}{discovered regions} \\
\cmidrule(lr){2-3}\cmidrule(lr){4-5}
profile & AUPRC & pointing & AUPRC & pointing \\
\midrule
flat (broadcast)   & \best{0.634} & 0.525 & 0.515 & 0.312 \\
occ-1              & 0.538 & \best{0.600} & 0.481 & 0.588 \\
occ-6              & 0.595 & 0.562 & \best{0.543} & \best{0.600} \\
occ-6 +f (default) & \runner{0.606} & 0.550 & \runner{0.542} & \best{0.600} \\
gradient           & 0.563 & 0.550 & 0.499 & \best{0.600} \\
gradient$\times$input & 0.523 & \best{0.600} & 0.467 & \best{0.600} \\
rank-align $0.5$   & 0.606 & 0.550 & 0.542 & \best{0.600} \\
\bottomrule
\end{tabular}
\end{table}

The invariant is exact: six arms spanning two perturbation families (structural occlusion and
finite-difference gradients) and three uniform-floor settings return the \emph{same} faithfulness to
three decimals. Only \texttt{rank-align}, which by design reorders regions, changes it---and it
changes it for the worse ($0.620\!\to\!0.561$, $0.582\!\to\!0.534$) while leaving localization
untouched, so we leave it off by default.

What the profile does control is a genuine trade, and its direction \emph{flips} with the region
regime. With ground-truth regions, the flat broadcast is best on area metrics (AUPRC $0.634$) and
worst at pointing ($0.525$): when the annotated span \emph{is} the ground truth, spreading credit
uniformly maximizes overlap and any sharpening can only lose it. With discovered regions the
smoothed occlusion profile wins both ($0.543$ AUPRC, $0.600$ pointing), because imprecise boundaries
leave real slack for the profile to recover. The honest summary is that within-region redistribution
earns its keep exactly when the segmentation does not already hand you the answer.

\paragraph{On real data, the trade is a precision/recall axis, not a single winner.}
Because the arms cannot move faithfulness, the sweep can be run the cheap way---one Owen pass per
instance, all six profiles scored off the same attribution---which is what makes it affordable on
real ECG. Table~\ref{tab:salreal} repeats it on ecg and ludb, and the synthetic story survives only
in part.

\begin{table*}[t]
\centering
\caption{\textbf{Within-region redistribution on real data.} Localization only: the profile cannot change faithfulness (the roll-up invariant), which the synth sweep confirms to three decimals. $n{=}80$ per ecg cell (8 instances $\times$ 5 tasks $\times$ 2 seeds), $n{=}16$ per ludb cell; $\pm$ is the between-seed SEM. AUPRC/AUR reward covering the annotated span, AUP/pointing reward hitting its peak---the two families disagree, and that disagreement is the result.}
\label{tab:salreal}
\footnotesize
\begin{tabular}{@{}lcccc cccc@{}}
\toprule
& \multicolumn{4}{c}{ecg} & \multicolumn{4}{c}{ludb} \\
\cmidrule(lr){2-5}\cmidrule(lr){6-9}
& \multicolumn{2}{c}{provided} & \multicolumn{2}{c}{discovered}
& \multicolumn{2}{c}{provided} & \multicolumn{2}{c}{discovered} \\
\cmidrule(lr){2-3}\cmidrule(lr){4-5}\cmidrule(lr){6-7}\cmidrule(lr){8-9}
profile & AUPRC & point & AUPRC & point & AUPRC & point & AUPRC & point \\
\midrule
flat (broadcast)      & \best{0.688} & 0.637 & \best{0.379} & 0.588 & \best{0.630} & 0.500 & 0.193 & 0.625 \\
occ-1                 & 0.406 & 0.625 & 0.280 & 0.562 & 0.316 & \best{0.625} & 0.197 & 0.562 \\
occ-6                 & 0.504 & \best{0.650} & 0.363 & 0.675 & 0.436 & \best{0.625} & \best{0.254} & \best{0.750} \\
occ-6 +f (default)    & \runner{0.594} & 0.637 & \runner{0.375} & \best{0.700} & \runner{0.499} & \best{0.625} & \runner{0.249} & \best{0.750} \\
gradient              & 0.512 & 0.613 & 0.269 & 0.487 & 0.424 & \best{0.625} & 0.177 & 0.688 \\
gradient$\times$input & 0.422 & 0.625 & 0.267 & 0.450 & 0.326 & \best{0.625} & 0.170 & 0.625 \\
\bottomrule
\end{tabular}
\end{table*}

The clean regime flip we saw on synth is really a \emph{precision-versus-recall} split that the
synthetic data was too small to separate. Across all four real cells, the occlusion profiles win the
precision family---AUP and pointing---and the flat broadcast wins the recall family---AUR and, where
the region support is handed over, AUPRC and IoU. This is what the two profiles are for: uniform
credit maximizes overlap with a span, a peaked profile maximizes agreement with a point.

What the region regime controls is how much of the recall advantage flat retains. With ground-truth
support on ecg it is large (AUPRC $0.688$ vs $0.594$); with discovered regions it collapses to
nothing ($0.379$ vs $0.375$); and on ludb's discovered regions it \emph{reverses}
($0.193$ vs $0.249$), matching synth. The mechanism is the same one the synthetic sweep suggested:
when the annotated span is the region, coverage comes free, and there is nothing for a profile to
recover, and when boundaries are approximate there is slack the profile can claim. The real data
just shows this is a matter of degree rather than a switch.

The default (\texttt{occ-6 +f}) is defensible on this evidence without being the winner of every
column. It is never worst in any cell; it is best or tied-best at pointing in the two discovered
cells---the regime that matters at deployment, where nobody hands you the regions---and it is
runner-up on AUPRC in all four. The pure-gradient profiles are the clear losers on real data
(worst or near-worst in seven of eight columns), which is worth stating: a finite-difference
sensitivity inside a region is a poor guide to where that region's \emph{coalitional} credit
belongs. We report the ludb cells as directional only---at $n{=}16$ the between-seed SEMs on AUPRC
reach $\pm0.10$.

\section{Swapping the generator, and asking whether it fits}\label{sec:genresults}
Section~\ref{sec:play} argues the generator is a component rather than a commitment, and that its adequacy can be checked directly instead of inferred from task metrics. Both claims are cheap to test, and neither is flattering in the way we expected.

\begin{table}[t]
\centering
\caption{\textbf{Three generators behind one game} (synth, $n{=}40$, $\pm$ between-seed SEM). The DDPM runs at its reduced budget ($P{=}16$, $Q{=}1$ vs.\ $128$/$3$), so its faithfulness is not budget-matched and is a lower bound.}
\label{tab:generators}
\footnotesize
\begin{tabular}{@{}lccc@{}}
\toprule
generator & faith$_{\text{on}}$ $\uparrow$ & faith$_{\text{off}}$ $\uparrow$ & loc-AUPRC \\
\midrule
GP, analytic (provided)   & \best{0.662}\,\tiny{$\pm$.025} & $-0.092$ & \best{0.585} \\
GP, analytic (discovered) & 0.510\,\tiny{$\pm$.010} & 0.017 & 0.574 \\
cVAE                      & 0.610\,\tiny{$\pm$.064} & 0.335 & 0.500 \\
DDPM$^{\ast}$             & 0.430\,\tiny{$\pm$.027} & \best{0.439} & 0.533 \\
\bottomrule
\end{tabular}
\end{table}

\textbf{Swappability holds, and there is a gradient behind it.} All three generators leave the game exactly complete, so the interface claim is met: the counterfactual moves and the accounting does not. What is more informative is the ordering. As the generator becomes more learned and less analytic, on-manifold faithfulness falls ($0.662\to0.610\to0.430$) while \emph{off}-manifold faithfulness rises monotonically ($-0.092\to0.335\to0.439$). The learned generators drift towards producing counterfactuals a splice-based operator recognizes, which is precisely the property the analytic GP was chosen to avoid. Part of the DDPM's on-manifold deficit is its $8\times$ smaller permutation budget and we do not read its $0.430$ as a converged number; the off-manifold \emph{ordering}, which is budget-insensitive in direction, is the part we rely on.

\begin{table}[t]
\centering
\caption{\textbf{Generator diagnostics} ($n{=}20$ per cell): marginal log-density per sample
($\uparrow$), PIT KS distance of the whitened residual from $\mathcal{N}(0,1)$ ($\downarrow$,
$0$ = honest uncertainty), and the two regeneration distances ($\downarrow$). Defined only for the
GP arms; the cVAE and DDPM have no tractable density and are reported as absent, not as zero.}
\label{tab:gencal}
\setlength{\tabcolsep}{4pt}
\footnotesize
\begin{tabular}{@{}llcccc@{}}
\toprule
 & regions & log-dens. & PIT-KS & stat-d & spec-d \\
\midrule
\multirow{2}{*}{synth}
 & provided   & $-17.8$ & 0.387 & 2.24 & 0.73 \\
 & discovered & $-21.5$ & 0.397 & 2.61 & 0.84 \\
\midrule
\multirow{2}{*}{ecg}
 & provided   & \best{$-12.4$} & \best{0.233} & \best{0.36} & \best{0.36} \\
 & discovered & $-67.0$ & 0.269 & 0.66 & 0.47 \\
\bottomrule
\end{tabular}
\end{table}

\textbf{The generator fits real ECG better than synthetic noise.} Per-sample log-density is highest
on ecg with annotated regions ($-12.4$) and lowest on synth ($-17.8$ to $-21.5$), and the
regeneration distances agree ($0.36$ vs.\ $2.24$). This inverts the intuition that synthetic data is
the easy case: synth's informative regions sit on i.i.d.\ background, which a smooth
boundary-conditioned GP is a poor model of, whereas a QRS complex is exactly the kind of correlated
excursion it represents well. It also supplies a mechanism for a result we previously only observed:
the exact game's large synth advantage on-manifold is measured against a generator that fits synth
\emph{worst}.

\textbf{Segmentation quality is visible in the generator, not just downstream.} On ecg, moving from
annotated to discovered regions costs $54$ nats per sample ($-12.4\to-67.0$)---a far larger swing
than anything in the synth column, and larger than the corresponding faithfulness gap would suggest.
Change-point boundaries place region edges where the GP's boundary conditioning is badly wrong. This
makes the log-density useful as a \emph{segmentation} diagnostic that needs no ground-truth mask, which
is exactly the setting where boundary $F_1$ is unavailable.

\textbf{The uncertainty is not honest, and we should say so.} PIT KS distances of $0.23$--$0.40$ are
large: if the generator's posterior were calibrated, the whitened residual $L^{-1}(x_r-\mu_r)$ would
be standard normal and this number would be near zero. It is not. The generator's \emph{mean} is
evidently good enough to drive the most faithful attributions in this paper while its \emph{error
bars} are miscalibrated---which is the whole point of reporting a diagnostic that task performance
cannot see. Two consequences follow. Any use of \method{} that reads the generator's variance
(confidence intervals on a counterfactual, say) is not currently supported by this evidence. And the
downstream faithfulness metrics are demonstrably insensitive to error-bar calibration, so passing
them is weaker evidence of a good generative model than it appears.

\section{Static covariates on one scale}\label{app:statics}
Section~\ref{app:impl} admits static covariates as singleton Owen players alongside the temporal
regions, and Table~\ref{tab:all_results} reports what they receive. This appendix gives the full
accounting, because the claim is otherwise only structural.

\textbf{The main tables hold the statics fixed.} In every run reported in the main text the observed
covariates are bound into the model, so each method explains the signal with the patient's context
held constant---which is what keeps the comparison against the eight signal-only baselines
like-for-like. The game reported here is the other one: sex and age are players, priced in the same
log-odds units as QRS. It is run on the same instances, the same region support, the same codec and
reference, and the same estimator budget, so the two are paired instance-by-instance.

\begin{table}[t]
\centering
\caption{\textbf{Static covariates as Owen players} (ludb $n{=}40$, mixed $n{=}30$; $\pm$ is the SEM over instances). \textbf{Bold} rows are static covariates. The remaining ones are temporal regions in the same game and the same units. On ludb, age out-scores every ECG region and is the highest-ranked player on $62.5\%$ of instances, while sex sits an order of magnitude below it. \texttt{mixed} is the control: its covariates have \emph{declared} roles, and the signed means recover them (shortcut and informative are $80\sigma$ and $8\sigma$ from zero, the distractor $1\sigma$).}
\label{tab:statics}
\footnotesize
\begin{tabular}{@{}llccc@{}}
\toprule
& Player & mean $\phi$ & mean $|\phi|$ & top-1 rate \\
\midrule
\multirow{6}{*}{\rotatebox{90}{ludb}}
 & \textbf{age} \tiny{(static)} & \best{1.265}\,\tiny{$\pm$.227} & \best{1.663} & \best{0.62} \\
 & \textbf{sex} \tiny{(static)} & 0.067\,\tiny{$\pm$.037} & 0.238 & 0.00 \\
 & QRS & \runner{0.891}\,\tiny{$\pm$.432} & \runner{1.430} & \runner{0.33} \\
 & T-wave & 0.141\,\tiny{$\pm$.089} & 0.452 & 0.05 \\
 & P-wave & 0.077\,\tiny{$\pm$.052} & 0.174 & 0.00 \\
 & \tiny{background spans (mean of 14)} & -- & 0.132 & -- \\
\midrule
\multirow{4}{*}{\rotatebox{90}{mixed}}
 & \textbf{class code} \tiny{(shortcut)} & \best{0.501}\,\tiny{$\pm$.006} & 0.501 & 0.00 \\
 & \textbf{mean} \tiny{(informative)} & 0.390\,\tiny{$\pm$.048} & 0.425 & 0.00 \\
 & \textbf{jitter} \tiny{(distractor)} & 0.063\,\tiny{$\pm$.064} & 0.253 & 0.00 \\
 & \tiny{injected + background regions} & -- & 2.078 & -- \\
\bottomrule
\end{tabular}
\end{table}

\begin{table}[t]
\centering
\caption{\textbf{With statics as players vs.\ held fixed}, paired per instance (ludb $n{=}40$,
mixed $n{=}30$; $^{*}$ = paired Wilcoxon $p<0.05$). Admitting the covariates is close to free in
faithfulness and localization; what it changes is the \emph{ranking}. On ludb the expected-region
hit rate is unmoved among signal players but falls to $0.325$ once age competes with QRS.}
\label{tab:staticabl}
\setlength{\tabcolsep}{4pt}
\footnotesize
\begin{tabular}{@{}llccc@{}}
\toprule
& & held fixed & as players & $\Delta$ \\
\midrule
\multirow{5}{*}{\rotatebox{90}{ludb}}
 & faith$_{\text{on}}$ $\uparrow$ & 0.598 & 0.597 & $-0.001$ \\
 & loc-AUPRC $\uparrow$ & 0.473 & 0.473 & $0.000$ \\
 & expected-region hit \tiny{(signal only)} & 0.575 & 0.625 & $+0.050$ \\
 & expected-region hit \tiny{(all players)} & 0.575 & 0.325 & $-0.250^{*}$ \\
 & $\Sigma|\phi|$ on signal regions & 5.901 & 5.983 & $+0.082$ \\
\midrule
\multirow{5}{*}{\rotatebox{90}{mixed}}
 & faith$_{\text{on}}$ $\uparrow$ & 0.786 & 0.788 & $+0.002$ \\
 & loc-AUPRC $\uparrow$ & 0.680 & 0.683 & $+0.003$ \\
 & expected-region hit \tiny{(signal only)} & 0.833 & 0.800 & $-0.033$ \\
 & expected-region hit \tiny{(all players)} & 0.833 & 0.800 & $-0.033$ \\
 & $\Sigma|\phi|$ on signal regions & 23.759 & 23.879 & $+0.120$ \\
\bottomrule
\end{tabular}
\end{table}

\begin{figure}[t]
\centering
\includegraphics[width=0.92\columnwidth]{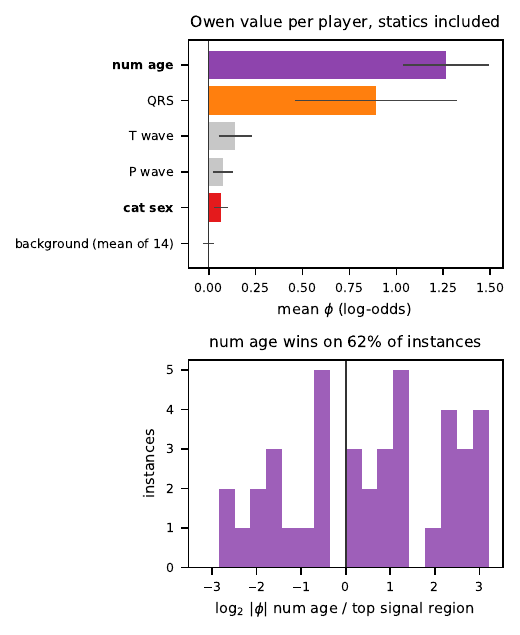}
\caption{\textbf{Age against the signal, LUDB LVH} ($n{=}40$). \textbf{Top:} mean Owen value per
player with the static variables in the game (bold = static, orange = the expected QRS region, unnamed
background spans pooled); age carries more mass than any ECG region, sex carries little.
\textbf{Bottom:} per instance, $\log_2$ of age's $|\phi|$ over the largest $|\phi|$ among all
temporal regions. Mass to the right of $0$ is an instance where the covariate outweighs the whole
signal---$62\%$ of them. The ratio is plotted rather than the raw pair because a single instance
reaches $|\phi|{=}14$ on its top region and would otherwise compress the rest into a corner.}
\label{fig:staticphi}
\end{figure}

\paragraph{Age dominates, sex does not, and completeness is untouched.}
Table~\ref{tab:statics} gives the accounting. Age's mean $|\phi|$ of $1.663$ exceeds the top
temporal region's $1.542$ and the QRS's $1.430$; the two covariates together take $24.8\%$ of the
total $|\phi|$, against a mean of $0.132$ for an unnamed background span. Sex at $0.238$ is where
\S\ref{app:setup} said it would be. The extra players do not perturb the accounting:
$\overline{|\rho|}=4.8\!\times\!10^{-8}$, i.e.\ $v(N)=f(x)$ to machine precision with the static
channel active. An independent, non-Shapley check agrees: flipping age alone to its cohort-median
reference and holding everything else observed moves the target log-odds by $+1.149$ on average,
the same sign and roughly the magnitude of its mean $\phi$ ($+1.265$).

\paragraph{The distractor control.}
On real data we have no ground truth for what a covariate \emph{should} receive, so the synthetic
\texttt{mixed} family carries the falsifiable version: its three covariates are injected with
declared roles. The signed means recover the declared ordering---shortcut $0.501\pm.006$,
informative $0.390\pm.048$, distractor $0.063\pm.064$---with only the distractor consistent with
zero. Note the contrast in \emph{share}: statics take $24.8\%$ of $|\phi|$ on ludb but only $4.9\%$
on mixed, where the injected temporal features are the whole signal. The static channel is not
biased towards attributing to statics; it reports what the model uses.

\paragraph{What it costs, and one caveat.}
Table~\ref{tab:staticabl} pairs the two games. Faithfulness and localization are statistically
unmoved on both families---the covariates are additional players, not a different value function
over the signal. The one significant movement is the expected-region hit rate under the joint
ranking, and it is the finding rather than a regression: on $25\%$ of instances age out-ranks the
annotated QRS. This is the quantitative form of the localization deficit LUDB shows throughout the
paper (\S\ref{app:discussion}). \emph{Caveat:} the held-fixed arm here is a re-run, not the frozen
row. It reproduces the frozen \texttt{ludb} faithfulness closely (faith$_{\text{on}}$ $0.598$ vs.\
$0.601$) but its loc-AUPRC is $0.473$ rather than the frozen CSV's $0.353$, because that CSV
predates the shipped occ-$w{=}6$ saliency default (\S\ref{sec:salabl}); $0.473$ is the comparable
number, and the $\Delta$ column is paired within this run either way.

\section{Amortized deployment for fast attribution}\label{app:amortized}
The exact GP game is faithful and exactly complete but slow. Because it is a \emph{value function}, the same attribution can be recovered by a learned network trained against it. This section emphasizes amortized deployment and the trade-off with faithfulness.

\subsection{Amortizing the game for deployment}\label{sec:amortized}
The exact estimator's cost is $|N|{+}1$ generator-and-model evaluations per permutation. We amortize it into a network that reproduces the same region attribution in one forward pass. We report two heads, because the difference between them is the finding of this appendix.

\paragraph{\amort{}: a FastSHAP density.} The first head is a convolutional map $g_\psi:\mathbb{R}^{\TT}\!\to\!\mathbb{R}^{\TT}$ emitting a per-timestep attribution density (pooled per region at explain time, so it is segmentation-agnostic), trained with the FastSHAP objective~\cite{jethani2021fastshap}: sampled coalitions with the additive-efficiency constraint enforced by projection given the measured $v(\emptyset)$ and $v(N)$. A convolutional density is translation-\emph{equivariant} and cannot express position-dependent attribution, so we add a positional input channel, without which the network converges to its target mean and learns nothing. Its training coalitions are realized with a cross-sample removal operator, and it is correspondingly strong off-manifold and only \emph{real-data}-competitive on-manifold---a deficit we first read as a training-operator mismatch and then measured (\S\ref{sec:distill}).

\paragraph{\amort{}: region tokens against the on-manifold game.} The deployed head replaces both the target and the architecture. It regresses the \emph{Owen values themselves}, computed by the exact on-manifold game of \S\ref{sec:define}, rather than coalition values under a splice operator. A dilated convolutional trunk over five input channels---signal, position, region-boundary indicator, own-region width, and a normalized occlusion-sensitivity profile that carries \emph{model evidence} into the input---is pooled per region into a token (mean/std/max, both endpoints, and eight geometry features), and the tokens are mixed by a small transformer encoder, so a region's attribution is a function of the other regions present rather than of its own span alone. Three heads read out shape ($\psi_r$), log-magnitude ($\log\sum_r|\phi_r|$) and $v(\emptyset)$ separately, which decouples ranking from scale; the training loss is a masked Pearson term on $\psi$ against the teacher $\phi$ (plus a small MSE, scale, $v(\emptyset)$, and within-region profile-KL term), so the objective is the ranking the faithfulness metric measures. Teachers are $400$--$1000$ signals per run at $\rho_{\text{holdout}} {=}0.75$--$0.99$.

Both heads keep the efficiency projection, so $\sum_r\phi_r=f(x)-v(\emptyset)$ holds \emph{exactly} at
inference and completeness is preserved by construction (measured $|\rho|<10^{-14}$ for both, all four
families). Because both are smooth functions of the input, their attribution is stable under input
\emph{time-shift}, where the exact estimator's fixed region-grid is most sensitive.

\subsection{Runtime and robustness}\label{sec:paid}
The exact game's main practical cost is runtime, and both heads pay it down (Table~\ref{tab:paid}).
\amort{} runs in $9$--$35$\,ms against $2.7$--$12.5$\,s---a $290$--$710\times$ speed-up---at a baseline-level effective-player count ($6$--$14$ vs.\ \method{}'s $27$--$53$), while remaining exactly complete. \amort{} is a further $3$--$5\times$ faster ($0.7$--$3.2\times10^{3}$ overall), and the gap is the price of the model-evidence channel: \amort{} runs an occlusion sweep over the signal before its forward pass, so it spends classifier calls that \amort{} does not. Section~\ref{sec:trade} highlights this difference.

Robustness was measured on the \amort{} head: both it and \method{} are jitter-stable ($\ge0.87$), and max-sensitivity \emph{improves} under amortization on the real families (ecg $0.76\!\to\!0.37$, ludb $0.22\!\to\!0.05$) but \emph{worsens} on synthetic ($0.68\!\to\!1.00$, $0.85\!\to\!1.07$). We do not carry that result over to \amort{}: its occlusion channel is a function of the classifier, not of the signal alone, so its perturbation response need not match \amort{}'s.

\begin{table}[t]
\centering
\caption{\textbf{\amort{} runtime} Method exactly complete and carry baseline-level effective-player counts; \amort{} gives back $\sim$$5\times$ of \amort{}'s speed-up for the model-evidence channel. Runtimes are \emph{medians}: each run's first explained instance carries a one-off $\sim$$1.6$\,s warm-up that inflates a mean over $20$--$100$ instances (mean-based speed-ups are $33$--$257\times$).}
\label{tab:paid}
\footnotesize
\begin{tabular}{@{}lccc cc@{}}
\toprule
& \multicolumn{3}{c}{runtime (s) $\downarrow$} & \multicolumn{2}{c}{eff.\ players} \\
\cmidrule(lr){2-4}\cmidrule(lr){5-6}
Family & \method & Amort. & Amort. & \method & Amort. \\
\midrule
synth & 4.53  & 0.014 & \best{0.006} & 27.1 & 6.1 \\
mixed & 2.72  & 0.009 & \best{0.003} & 28.9 & 5.6 \\
ecg   & 12.55 & 0.035 & \best{0.006} & 45.7 & 9.0 \\
ludb  & 8.61  & 0.012 & \best{0.003} & 52.9 & 13.9 \\
\bottomrule
\end{tabular}
\end{table}

\subsection{On-manifold faithfulness tradeoff}\label{sec:trade}
On-manifold faithfulness is where the exact game is unbeatable and where a surrogate performs worse (Table~\ref{tab:trade}). \method{} remains the best across all four families. But the difference depends almost entirely on the head. \amort{} beats the best baseline on all four datasets, including synth.

What remains is a real but much narrower gap to the exact game. On ecg it is $0.052$ ($0.776$ vs.\ $0.828$)---significant at $q{=}8\times10^{-6}$ but small enough that the ordering, not the level, is what a $250\times$ cheaper explainer gives up. On ludb the gap ($0.124$) is not significant ($q{=}0.12$, $n{=}20$): at this sample size, the deployed head is statistically indistinguishable from the game it was distilled from. Synth and mixed keep a clear, significant gap ($0.316$ and $0.228$, $q<10^{-5}$), so the exact GP game remains the on-manifold reference and the recommendation is unchanged in kind---only in degree.

Localization moves the same way and further. \amort{} improves loc-AUPRC over \amort{} on all four families and beats the exact game on both real ones: ecg $0.648$ vs.\ $0.587$ ($q{=}1.6\times10^{-3}$) and ludb $0.633$ vs.\ $0.502$ ($q{=}0.09$), the ecg figure also clearing the best baseline in the full eight-method run ($0.607$). It is not a uniform win: on mixed it still trails \method{} ($0.518$ vs.\ $0.648$), and its \emph{pointing} score is worse on mixed ($0.10$ vs.\ $0.60$) and ecg ($0.55$ vs.\ $0.64$). It distributes mass over the annotated region better than the exact game, but places its single peak worse.

\begin{table}[t]
\centering
\caption{\textbf{The tradeoff} (\texttt{faithfulness\_corr\_on}, mean\,$\pm$\,SEM,
$n_{\text{amo}}$). The exact game still leads, but the deployed head now beats the best baseline across all datasets. $^{\dagger}$gap to \method{} not significant (BH-adjusted Wilcoxon $q{=}0.12$); every other
\amort{}$-$\method{} gap has $q<10^{-4}$. Baselines are the four-method $n_{\text{amo}}$ subset (\S\ref{app:setup}); \amort{} also clears the full eight-method leader on all four families.}
\label{tab:trade}
\footnotesize
\begin{tabular}{@{}lccc@{}}
\toprule
Family & \method{} (exact) & \amort{} & best baseline \\
\midrule
synth & \best{$0.872$}\,\tiny{$\pm$.024} & $0.556$\,\tiny{$\pm$.071}           & $-0.114$ \\
mixed & \best{$0.921$}\,\tiny{$\pm$.011} & $0.694$\,\tiny{$\pm$.047}           & $0.595$ \\
ecg   & \best{$0.828$}\,\tiny{$\pm$.016} & $0.776$\,\tiny{$\pm$.021}            & $0.646$ \\
ludb  & \best{$0.657$}\,\tiny{$\pm$.073} & $0.533^{\dagger}$\,\tiny{$\pm$.076} & $0.463$ \\
\bottomrule
\end{tabular}
\end{table}

\subsection{Is the surrogate's negative a training-operator mismatch?}\label{sec:distill}
There is an obvious suspect for the synthetic negatives, and it is worth naming before defending
anything: \amort{} regresses coalition values built by \emph{cross-sample splicing}---absent spans
filled from another signal---and is then scored under the on-manifold operator. Fitting one removal
operator and grading against another is a semantic mismatch, and it would explain the sign.

We tested it directly rather than arguing about it. We added a second distillation target that
evaluates coalitions through the on-manifold GP game itself---the exact value function
$v(S)$ of \S\ref{sec:define}, the same one \method{} is computed against---and left the FastSHAP
machinery downstream untouched, so the training operator is the only thing that moves. Both heads
were trained on identical corpora ($200$ signals, $24$ coalitions each) and evaluated on identical
instances and seeds (Table~\ref{tab:distill}).

\begin{table}[t]
\centering
\caption{\textbf{Distillation target ablation} ($n{=}80$ per cell, $20$ instances $\times$ 2 splits
$\times$ 2 seeds; $\pm$ is the between-seed SEM). Training \amort{} against the on-manifold value
function instead of cross-sample splices does \emph{not} recover on-manifold faithfulness. Paired
\method{} on the same instances is shown for scale.}
\label{tab:distill}
\setlength{\tabcolsep}{5pt}
\footnotesize
\begin{tabular}{@{}llcc@{}}
\toprule
 & distillation target & faith$_{\text{on}}$ $\uparrow$ & loc-AUPRC $\uparrow$ \\
\midrule
\multirow{3}{*}{synth}
 & cross-sample (default) & $-0.183$\,\tiny{$\pm$.071} & $0.415$\,\tiny{$\pm$.014} \\
 & on-manifold            & $-0.195$\,\tiny{$\pm$.031} & \best{$0.452$}\,\tiny{$\pm$.026} \\
 & \method{} (exact, paired) & \best{$0.620$} & \best{$0.606$} \\
\midrule
\multirow{3}{*}{mixed}
 & cross-sample (default) & $0.003$\,\tiny{$\pm$.084} & $0.419$\,\tiny{$\pm$.001} \\
 & on-manifold            & $0.014$\,\tiny{$\pm$.113} & $0.421$\,\tiny{$\pm$.004} \\
 & \method{} (exact, paired) & \best{$0.784$} & \best{$0.624$} \\
\bottomrule
\end{tabular}
\end{table}

\textbf{The mismatch is not the explanation.} Matching the training operator to the evaluation operator moves faithfulness by $-0.012$ on synth and $+0.011$ on mixed. Both are far inside a single between-seed SEM, and on synth in the wrong direction. The negative survives its most obvious confound. What the on-manifold target \emph{does} buy is localization on synth ($0.415\!\to\!0.452$), i.e.\ a better-placed explanation that is no more faithful: the head learns where the informative region is without learning how the model responds when that region is regenerated on-manifold.

We therefore retract the operator-mismatch reading and state that the residual explanation is the narrower claim supported by the data. A FastSHAP-style head is a \emph{single} amortized map from signal to attribution; the on-manifold value function it is asked to imitate is a function of the region generator as well as the classifier, and on low-signal synthetic data the classifier's response to a regenerated region is close to noise. There is little stable structure for a single forward pass to capture, regardless of which operator produced the targets. On real data---where regenerating a QRS actually changes the decision in a repeatable way---the same architecture and the same training budget succeed (ecg $0.556$, ludb $0.381$). This is a sharper statement than the one it replaces, and it was cheap to obtain: one extra target operator, four benchmark runs.

\section{Extended Discussion}\label{app:discussion}

\method{} is the method to reach for when the question is which regions the model's decision is most sensitive to, on-distribution and with a complete accounting: it is the most faithful on-manifold method on every family and exactly complete. \amort{} is the method to deploy: on real, structured data it recovers this at a small fraction of the cost, at a faithfulness price that is modest on real data and disqualifying on low-signal synthetic data.

\paragraph{Three lessons that generalize.} First, the costs of an exact on-manifold game are the solver's, not the game's: amortizing against the same value function pays down runtime without changing what is being explained (Section~\ref{sec:exp:deploy}). Second, faithfulness is operator-relative; reporting both on- and off-manifold operators for every method is the only way to avoid teaching-to-the-test, and it is precisely what exposes the surrogate's on-manifold limit (Section~\ref{sec:exp:fair}). Third, localization against a fixed anatomical prior is only as meaningful as that prior is to the model: the LUDB LVH case shows a faithful explanation can correctly point away from the annotated region when the model relies on static or un-annotated evidence, measurably so, since admitting age as a player takes top rank from the annotated QRS on a quarter of instances while leaving faithfulness unchanged (Section~\ref{sec:exp:covariates}).

\paragraph{Limits that are structural.} Three of our limits are structural, and each was reached by testing a claim we had already made in print. The generator's mean is good enough to drive the most faithful attributions here while its \emph{error bars} are not calibrated, so faithfulness metrics cannot certify a generative model (Section~\ref{sec:exp:generator}). Matching an amortized head's training operator to its evaluation operator does not rescue it on low-signal data, so the surrogate's deficit is not a semantic mismatch. And a conditional reference, however strongly its features co-vary, cannot alter a region-level attribution at all, because Owen block values only ever see whole-block coalitions (Section~\ref{sec:exp:robust}). Testing claims we had already committed to in print is the habit we would most like to see carried forward.


\end{document}